\documentclass[10pt,twocolumn,letterpaper]{article}
\usepackage[titletoc,toc,title]{appendix}
\usepackage{cvpr}
\usepackage{times}
\usepackage{epsfig}
\usepackage{graphicx}
\usepackage{amsmath}
\usepackage{amssymb}
\usepackage{subfig}
\usepackage{float}

\usepackage[breaklinks=true,bookmarks=false]{hyperref}

\cvprfinalcopy 

\def\cvprPaperID{3247} 

\ifcvprfinal\fi
\begin{document}

\title{Deformable GANs for Pose-based Human Image Generation}

\author{Aliaksandr Siarohin$^1$, Enver Sangineto$^1$, St{\'e}phane Lathuili{\`e}re$^2$, and Nicu Sebe$^1$\\
$^1$DISI, University of Trento, Italy, \hspace{\columnsep} $^2$ Inria Grenoble Rhone-Alpes, France\\
{\tt\small \{aliaksandr.siarohin,enver.sangineto,niculae.sebe\}@unitn.it}, {\tt\small stephane.lathuiliere@inria.fr}
}

\maketitle
\thispagestyle{empty}

\begin{abstract}
In this paper we address the problem of generating person images   conditioned on a given pose. 
Specifically, given an image of a person and a target pose, we synthesize a new image of that person in the novel pose. In order to deal with pixel-to-pixel misalignments caused by the pose differences, we introduce {\em deformable skip connections} in  the  generator  of our Generative Adversarial Network. Moreover, a nearest-neighbour loss is proposed instead of the common $L_1$ and $L_2$ losses in order to match the details of the generated image with the target image.
We test  our approach using  photos of persons in different poses and we compare our method with previous work in this area showing state-of-the-art results in two  benchmarks.
Our method can be applied to the wider field of deformable object generation, provided that the pose of the articulated object can be extracted using a keypoint detector.
\end{abstract}

\section{Introduction}
\label{Introduction}

In this paper we deal with the problem of generating images where the foreground object changes because of a viewpoint variation or a deformable motion, such as the articulated human body. 
Specifically, inspired by Ma et al. \cite{ma2017pose}, our goal is to generate a human image  conditioned on two different variables: (1) the appearance of a specific person in a given image and (2) the pose 
of the same 
person in another image.
The task our networks 
need to solve is to preserve the appearance details (e.g., the texture) contained in  the first variable while performing a deformation on the structure of the foreground object according to the second variable.
We focus on the human body which is an articulated ``object'', important for many applications (e.g., computer-graphics based manipulations or re-identification dataset synthesis). However, our approach can be used with other deformable objects such as human faces or animal bodies, provided that a significant number of keypoints can be automatically extracted from the object of interest in order to represent its pose.

\begin{figure}[t!]
\centering
\subfloat[Aligned task]{\includegraphics[height=0.6\columnwidth]{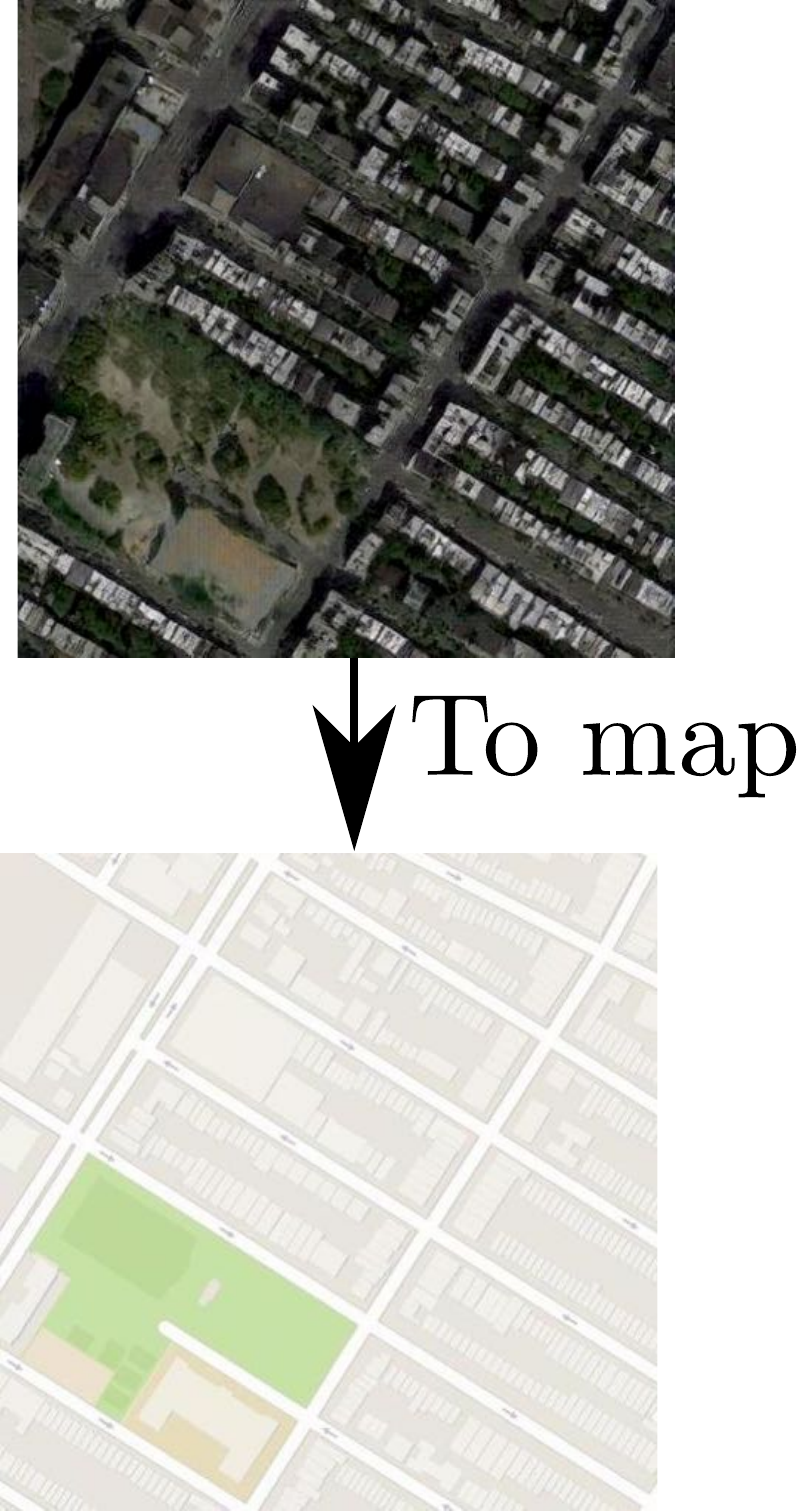}\label{fig:teaserAligned}}
\subfloat[Unaligned task]{\includegraphics[height=0.6\columnwidth]{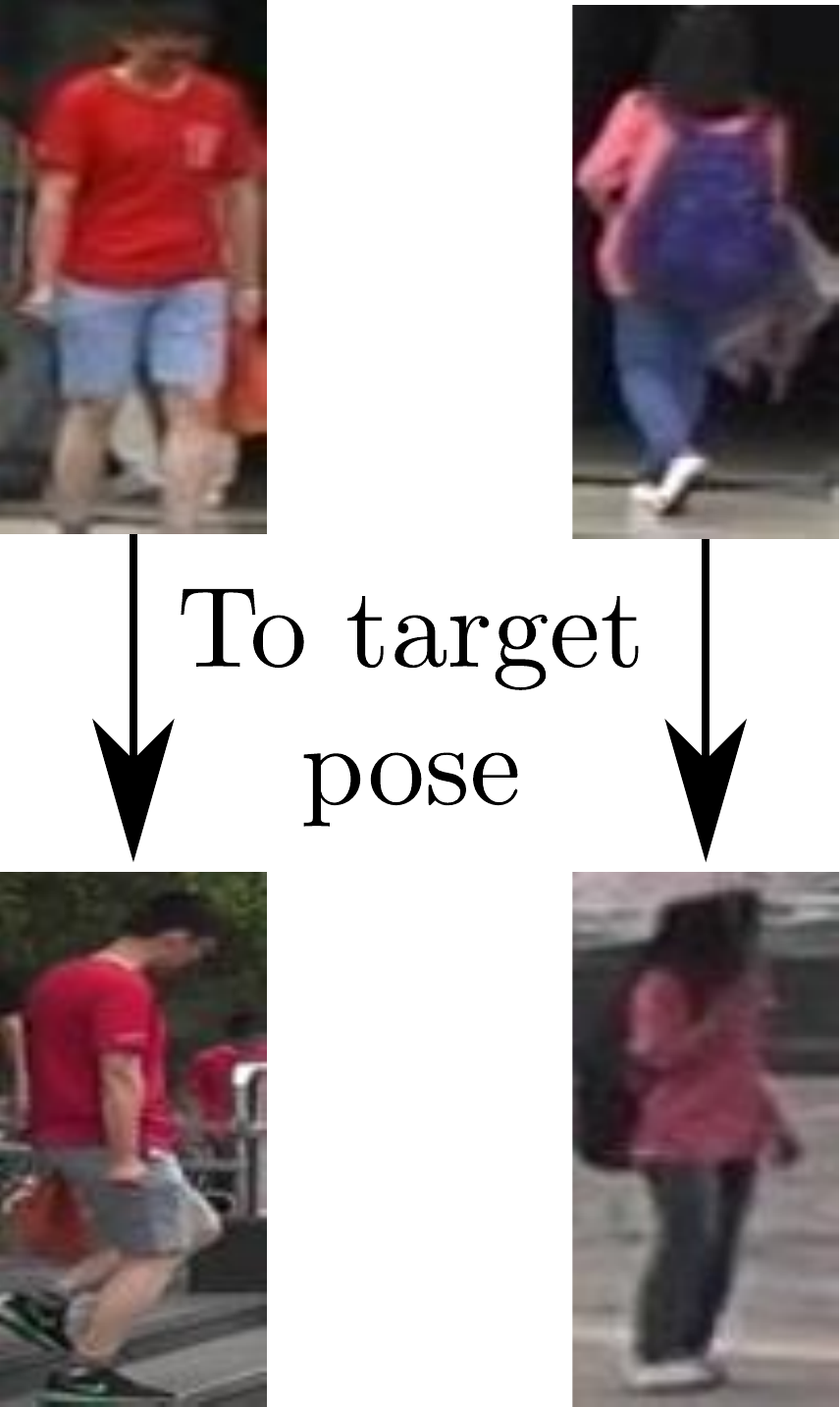}\label{fig:teaserUnaligned}}
\caption{(a) A typical ``rigid'' scene generation task, where the conditioning and the output image local structure is well aligned. (b) In a deformable-object  generation task, the input and output are not spatially aligned.}
\vspace{-0.5cm}
\label{fig:teaser}
\end{figure}

Pose-based human-being image  generation is motivated by the interest in synthesizing videos \cite{Walker2017pose} with non-trivial human movements or in generating rare poses  for human pose estimation \cite{Cao} or re-identification \cite{Zheng_2017_ICCV} training datasets. 
However, most of the recently proposed, deep-network based generative approaches, such as  Generative Adversarial Networks (GANs) \cite{goodfellow2014generative} or Variational Autoencoders (VAEs) 
 \cite{kingma2013auto} do not explicitly deal with the problem of articulated-object generation.
 Common conditional methods (e.g., conditional GANs or conditional VAEs) can synthesize images whose appearances depend on some conditioning variables (e.g., a label or another image). For instance, Isola et al. \cite{pix2pix2016} recently proposed an ``image-to-image translation'' framework, in which an input image $x$ is transformed into a second image $y$ represented in another ``channel'' (see Fig.~\ref{fig:teaserAligned}).
However, most of these methods have problems when dealing with large spatial deformations between the conditioning and the target image. 
For instance, 
the U-Net architecture used by  Isola et al. \cite{pix2pix2016} is based on {\em skip connections} which help preserving local information between $x$ and $y$. 
Specifically, skip connections are used to copy and then concatenate the feature maps of the generator ``encoder'' (where information is downsampled using convolutional layers) to the generator ``decoder'' (containing the upconvolutional layers).
However, the assumption used in \cite{pix2pix2016} is that $x$ and $y$ are roughly aligned with each other and they represent the same underlying structure. This assumption is violated 
when the foreground object in $y$ undergoes to large spatial deformations with respect to $x$ (see Fig.~\ref{fig:teaserUnaligned}). As shown in \cite{ma2017pose},
 skip connections  
cannot reliably cope with misalignments between the two poses.

Ma et al. \cite{ma2017pose} propose to alleviate this problem using a two-stage generation approach. In the first stage a U-Net generator is trained using a masked $L_1$ loss in order to produce an intermediate image conditioned on the target pose. In the second stage, a second U-Net based generator is trained using also an adversarial loss in order to generate an appearance difference map which brings the  intermediate image closer to the appearance of the conditioning image.
In contrast, the GAN-based method we propose in this paper  is end-to-end trained
  by explicitly taking into account  pose-related spatial deformations. More specifically, we propose {\em deformable skip connections} which ``move'' local information according to the structural deformations represented in the  conditioning variables. These layers are used  in our U-Net based generator. 
In order to  move information according to a specific spatial deformation,  we decompose the overall  deformation by means of a set of 
local affine  transformations involving subsets of joints, 
then we deform the convolutional feature maps of the encoder  according to these  transformations and we use common skip connections to transfer the transformed tensors to the decoder's fusion layers.
Moreover, we also propose to use a {\em nearest-neighbour 
loss} as a replacement of common pixel-to-pixel losses (such as, e.g., $L_1$ or $L_2$ losses) commonly used in conditional generative approaches. This loss proved to be helpful in generating local information (e.g., texture) similar to the target image which is not penalized because of small spatial misalignments.

We test our approach using the benchmarks and the evaluation protocols proposed in \cite{ma2017pose} obtaining higher qualitative and quantitative results in all the datasets.
Although tested on the specific human-body problem, our approach makes few human-related assumptions and can be easily extended to other domains involving the generation of  highly deformable objects.
Our code   and our trained models
are publicly available\footnote{\url{https://github.com/AliaksandrSiarohin/pose-gan}}.

\section{Related work}
\label{Related}

Most common deep-network-based
 approaches for visual content generation can be categorized as either Variational Autoencoders (VAEs) \cite{kingma2013auto} or Generative Adversarial Networks (GANs) \cite{goodfellow2014generative}. VAEs  are based on probabilistic graphical models and are trained by maximizing a lower bound of the corresponding data likelihood. GANs are based on two networks, a generator and a discriminator, which are trained simultaneously such that the generator tries to ``fool'' the discriminator and the discriminator learns how to distinguish between real and fake images.

 Isola et al. \cite{pix2pix2016} propose a  conditional GAN framework for image-to-image translation problems,
 where a given scene representation is ``translated'' into another representation.
The main assumption behind this framework is that there exits a spatial correspondence between the low-level information of the conditioning and the output image. 
VAEs and GANs are combined in \cite{ZhaoWCLF17} to generate realistic-looking multi-view clothes images from  a single-view
input image. The target view is filled to the model via a viewpoint label as \emph{front} or \emph{left side} and a two-stage approach is adopted: pose integration and image refinement.
Adopting a similar pipeline, Lassner et al. \cite{LassnerPG17} generate images of people with different clothes in a given pose. 
This approach is based on a costly annotation (fine-grained segmentation with 18 clothing labels) and a complex  3D pose representation. 

Ma et al. \cite{ma2017pose} propose  a more general approach  which allows to synthesize person images in any arbitrary pose. Similarly to our proposal, the input of their  model is a conditioning image of the person and a target new pose defined by 18 joint locations. The target pose is described by means of binary maps where small circles represent the joint locations. Similarly to \cite{LassnerPG17,ZhaoWCLF17},
the generation process is split in two different stages: pose generation and texture refinement. In contrast, in this paper we show that a single-stage approach, trained end-to-end, can be used for the same task obtaining higher qualitative results.

Jaderberg et al. \cite{jaderberg2015spatial}  propose a
 spatial transformer layer, which learns how to transform a feature map in a ``canonical'' view, conditioned on the feature map
itself.
However only a global,  parametric  transformation can be learned (e.g., a global affine transformation), while in this paper we deal with  non-parametric deformations of articulated objects which cannot be described by means of a unique global affine transformation.

Generally speaking,
  U-Net based architectures are frequently adopted for
pose-based person-image generation tasks \cite{LassnerPG17,ma2017pose,Walker2017pose,ZhaoWCLF17}. However, common U-Net skip connections  are not well-designed for large spatial deformations  because local information in the input and in the output images is not  aligned (Fig.~\ref{fig:teaser}).
In contrast, we propose deformable skip connections to deal with this misalignment problem and ``shuttle'' local information from the encoder to the decoder driven by the specific pose difference.
In this way, differently from previous work, we are able to simultaneously  generate
the overall pose and the texture-level refinement.
 
 Finally, our nearest-neighbour loss is similar to the
 perceptual loss proposed in \cite{DBLP:conf/eccv/JohnsonAF16} and to the 
  style-transfer spatial-analogy approach recently proposed in \cite{DBLP:journals/tog/LiaoYYHK17}. However, 
the perceptual loss,  based on an element-by-element difference computed in the feature map of an external classifier \cite{DBLP:conf/eccv/JohnsonAF16}, does not take into account spatial misalignments. On the other hand,   
  the patch-based similarity, adopted in \cite{DBLP:journals/tog/LiaoYYHK17} to compute a dense feature correspondence, is very computationally expensive and it is not used as a loss.

\section{The network architectures}
\label{architectures}

\begin{figure*}[t]\centering
\includegraphics[width=0.75\linewidth]{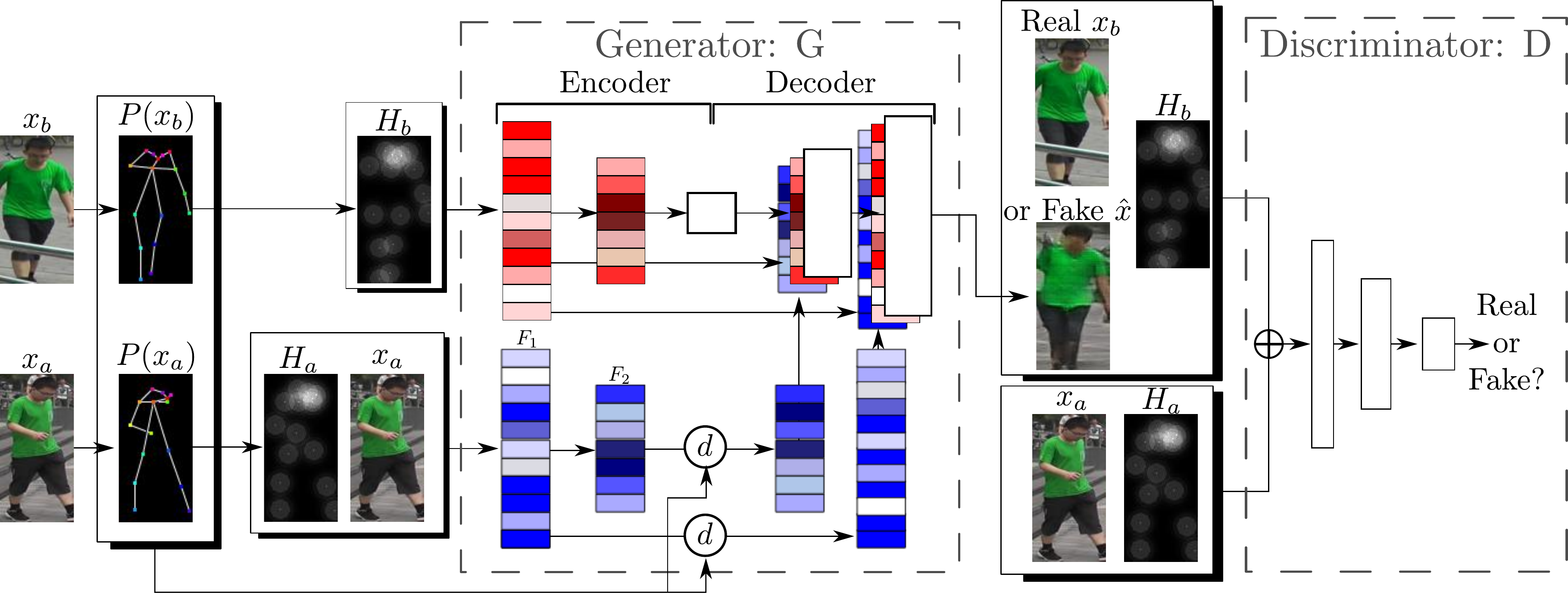}
\caption{A schematic representation of our network architectures. For the sake of clarity, in this figure we depict $P(\cdot)$ as a skeleton and each tensor $H$ as the average of its component matrices $H_j$ ($1 \leq j \leq k$). The white rectangles in the decoder represent the feature maps directly obtained using up-convolutional filters applied to the previous-layer maps. The reddish rectangles represent the feature maps ``shuttled'' by the skip connections from the $H_b$ stream. Finally, blueish rectangles represent the deformed tensors $d(F)$ ``shuttled'' by the deformable skip connections from the $(x_a,H_a)$ stream.}
\vspace{-0.5cm}
\label{fig:pipeline}
\end{figure*}

In this section we describe the architectures of our generator ($G$) and discriminator ($D$)  and the proposed deformable skip connections.  We first introduce  some notation. At testing time our task, similarly to \cite{ma2017pose}, consists in generating an image $\hat{x}$ showing a person whose appearance (e.g., clothes, etc.) is similar to an input, conditioning image $x_a$ but with a body pose similar to $P(x_b)$, where $x_b$ is a different image of the same person and $P(x) = (\mathbf{p}_1, ... \mathbf{p}_k)$ is a sequence of $k$ 2D points describing the locations of the human-body joints in $x$. 
In order to allow  a fair comparison with \cite{ma2017pose},
we use the same number  of joints ($k = 18$)
and we extract  $P()$ using the same Human Pose Estimator (HPE)  \cite{Cao} used in \cite{ma2017pose}. Note that this HPE is used both at testing and at training time, meaning that we do not use manually-annotated poses and the so extracted joint locations may have some localization errors or missing detections/false positives.

At training time we use a dataset  ${\cal X} = \{ (x_a^{(i)}, x_b^{(i)}) \}_{i=1,...,N}$ containing pairs of conditioning-target images of the same person in different poses.
For each pair $(x_a, x_b)$, a conditioning and a target pose $P(x_a)$ and $P(x_b)$ is extracted from the corresponding image and represented using two  tensors $H_a = H(P(x_a))$ and $H_b = H(P(x_b))$, each composed of $k$ heat maps, where  $H_j$ ($1 \leq j \leq k$) is a 2D matrix of the same dimension as the original image. If $\mathbf{p}_j$ is the j-th joint location, then:

\begin{equation}
H_j(\mathbf{p}) = exp\left(-\frac{\lVert \mathbf{p}- \mathbf{p}_j \rVert}{\sigma^2}\right),
\end{equation}

\noindent
with $\sigma = 6$ pixels (chosen with cross-validation).
Using
 blurring instead of a binary map is useful to provide widespread information about the location $\mathbf{p}_j$.

The generator $G$ is fed with: (1) a noise vector $z$, 
 drawn from a noise distribution ${\cal Z}$ and implicitly provided using dropout \cite{pix2pix2016} 
and (2) the triplet $(x_a, H_a, H_b)$. Note that, at testing time, the target pose is known, thus $H(P(x_b))$ can be computed.
Note also that the joint locations in $x_a$ and  $H_a$ are spatially aligned (by construction), while in $H_b$ they are different.
Hence, differently from \cite{ma2017pose,pix2pix2016}, $H_b$ is not concatenated with the other input tensors. Indeed the convolutional-layer units in the encoder part of $G$ have a small receptive field which cannot capture large spatial displacements. For instance, a large movement of a body limb in $x_b$ with respect to $x_a$, is represented in different locations in $x_a$ and $H_b$ which may be too far apart  from each other to be captured by the receptive field of the convolutional  units. This is emphasized in the first layers of the encoder, which represent low-level information. Therefore, the convolutional filters cannot simultaneously process texture-level information (from $x_a$) and the corresponding pose information (from $H_b$).

For this reason we independently process  $x_a$ and $H_a$ from $H_b$ in the encoder. 
Specifically, $x_a$ and $H_a$ are concatenated and processed
using a convolutional stream of the encoder while  $H_b$ is processed by means of  a second convolutional stream,  without sharing the weights 
(Fig.~\ref{fig:pipeline}).
The feature maps of the first stream are then fused with the layer-specific  feature maps of the second stream in the decoder  after a pose-driven spatial deformation performed by our deformable skip connections (see Sec.~\ref{skip-connections}).

Our discriminator network is based on the conditional, fully-convolutional 
discriminator proposed by
Isola et al. \cite{pix2pix2016}. In our case, $D$ takes as input 4 tensors: $(x_a, H_a, y, H_b)$, where either  $y =  x_b$ or $y =  \hat{x} = G(z, x_a, H_a, H_b)$ (see Fig.~\ref{fig:pipeline}).
These four tensors are concatenated  and then given as input to $D$. The  discriminator's output is a  scalar value indicating its confidence on the fact that  $y$ is a real image.

\subsection{Deformable skip connections}
\label{skip-connections}

As mentioned above  and similarly to  \cite{pix2pix2016},
the goal of the deformable skip connections is to ``shuttle'' local information from the encoder to the decoder part of $G$. The local information to be transferred is, generally speaking, contained in a tensor $F$, which represents the feature map activations of a given convolutional layer of the encoder.
However, differently from \cite{pix2pix2016}, we need to ``pick''
the information to shuttle taking into account  the object-shape deformation which is  described by the difference between $P(x_a)$ and $P(x_b)$.
To do so, we decompose the global deformation in a set of local affine transformations, defined using subsets of joints in $P(x_a)$ and $P(x_b)$.
Using these affine transformations and local masks constructed using the specific joints, we deform the content of $F$ and then we use common skip connections
to copy the transformed tensor and concatenate it with the corresponding tensor in the destination layer (see  Fig.~\ref{fig:pipeline}).
Below we describe in more detail the whole pipeline.

\begin{figure}[t!]\centering
\includegraphics[angle=-90,width=0.9\linewidth]{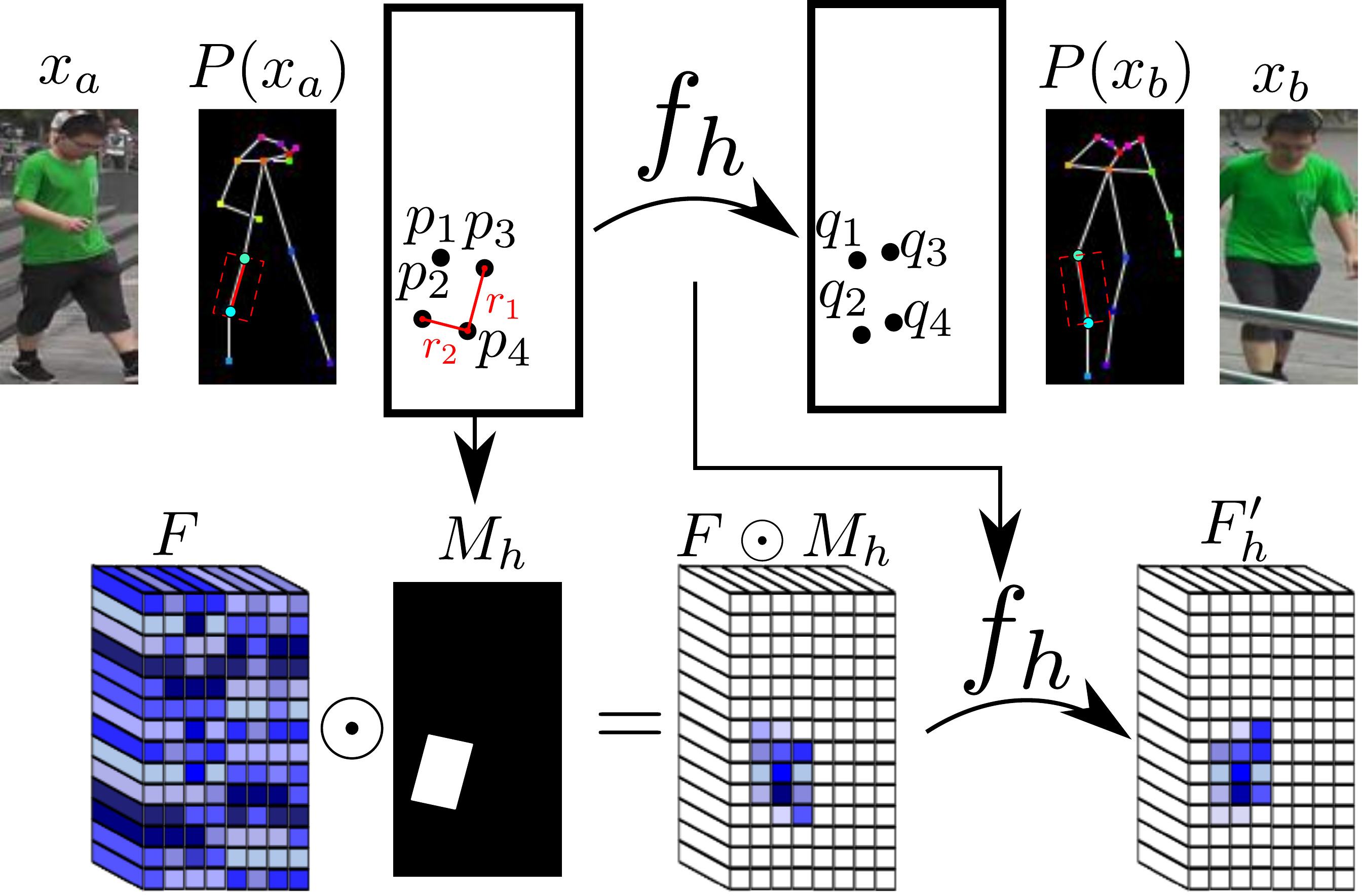}
\caption{For each specific body part, an  affine transformation $f_h$  is computed. This transformation is  used to ``move'' the feature-map content corresponding to that body part.}
\label{fig:affinepipeline}
\end{figure}

{\bf Decomposing an articulated body in a set of rigid sub-parts.} 
The human body is an articulated ``object'' which can be roughly decomposed into a set of rigid sub-parts. We chose 10 sub-parts: the head, the torso, the left/right upper/lower arm and the  left/right upper/lower leg. Each of them corresponds to a subset of the 18 joints defined by the HPE  \cite{Cao} we use for extracting $P()$. 
Using these  joint locations we can define rectangular regions which enclose the specific body part.
In case of  the head, the  region is simply chosen to be the axis-aligned enclosing rectangle of all the corresponding joints. For the torso, which is the largest area, we use a region which includes  the whole image, in such a way to shuttle  texture information for the background pixels.
Concerning the body limbs, each limb corresponds to only 2 joints. In this case we define a region to be a rotated rectangle whose major axis ($r_1$) corresponds to the line between these two joints, while the minor axis ($r_2$) is orthogonal to $r_1$ and with a length equal to one third of the mean of the torso's diagonals (this value is used for all the limbs). In Fig.~\ref{fig:affinepipeline} we show an example.
 Let $R_h^a = \{ \mathbf{p}_1, ..., \mathbf{p}_4 \}$ be the set of the 4 rectangle corners in $x_a$ defining the $h$-th body region ($1 \leq h \leq 10$).
 Note that these 4 corner points are not joint locations.
 Using $R_h^a$   we can compute a binary mask  $M_h(\mathbf{p})$ which is zero everywhere except those points $\mathbf{p}$ lying inside  $R_h^a$.
Moreover, let $R_h^b = \{ \mathbf{q}_1, ..., \mathbf{q}_4 \}$ be the corresponding rectangular region in $x_b$. Matching the points in $R_h^a$ with the corresponding points in $R_h^b$ we can compute the parameters of a body-part specific affine transformation (see below).
In  either  $x_a$ or $x_b$, some of the body regions can be occluded, truncated by the image borders or simply miss-detected by the HPE. In this case we leave  the corresponding region $R_h$ empty and the $h$-th affine transform is not computed (see below). 

Note  that our body-region definition  is the only human-specific part of the proposed approach. However, similar regions can be easily defined using the joints of other articulated objects such as those representing an animal body  or a human face.

{\bf Computing a set of affine transformations.} 
During the forward pass (i.e., both at training and at testing time) we decompose the global deformation of the conditioning pose with respect to the target pose by means of a set of local affine transformations, one per body region. Specifically, given $R_h^a$ in $x_a$ and $R_h^b$ in $x_b$ (see above), we compute the 6  parameters $\mathbf{k}_h$ of an affine transformation $f_h(\cdot ; \mathbf{k}_h)$
using Least Squares Error:

\begin{equation}
\min_{\mathbf{k}_h} \sum_{\mathbf{p}_j \in R_h^a, \mathbf{q}_j \in R_h^b} || \mathbf{q}_j - f_h(\mathbf{p}_j ; \mathbf{k}_h) ||^2_2
\end{equation}

The parameter vector 
$\mathbf{k}_h$ is computed using the original image resolution of $x_a$ and $x_b$ and then adapted to the specific resolution of each involved  feature map $F$.
Similarly, we compute scaled versions of each $M_h$.
In case either $R_h^a$ or $R_h^b$ is empty (i.e., when any of the specific body-region joints has not been detected using the HPE, see above), then we simply set   $M_h$ to be a matrix with all elements equal to 0 ($f_h$ is not computed).

 Note that  $(f_h(), M_h)$ and their lower-resolution variants need to be computed 
only once  per each pair of real images $(x_a, x_b) \in {\cal X}$ and, in case of the training phase, this is can be done before starting training the networks (but in our current implementation this is done on the fly).

{\bf Combining affine transformations to approximate the  object deformation.} 
Once $(f_h(), M_h)$, $h= 1, ...,10$ are computed for the specific spatial resolution of a given tensor $F$, the latter can be transformed in order to approximate the global pose-dependent deformation.
Specifically, we first compute for each $h$:

\begin{equation}
F'_h = f_h (F \odot M_h),
\end{equation}

\noindent
where $\odot$ is a point-wise multiplication
and $f_h(F(\mathbf{p}))$ is used to ``move'' all the channel values of $F$ corresponding to point 
$\mathbf{p}$.
Finally, we merge the resulting  tensors using:

\begin{equation}
\label{eq.d-F}
d(F(\mathbf{p},c)) = max_{h = 1, ..., 10} F'_h(\mathbf{p},c),
\end{equation}

\noindent
where $c$ is a specific channel.
The rationale behind Eq.~\ref{eq.d-F} is that, when two body regions partially overlap each other, the final deformed tensor $d(F)$ is obtained by picking the  maximum-activation values. Preliminary experiments performed using  average pooling  led to slightly worse results.

\section{Training}
\label{Training}

$D$ and $G$ are trained using a combination of a standard conditional adversarial loss ${\cal L}_{cGAN}$ with our proposed nearest-neighbour loss ${\cal L}_{NN}$. Specifically, in our case 
${\cal L}_{cGAN}$ is given by:

\begin{equation}
\label{eq.GAN-loss}
\begin{array}{cc}
{\cal L}_{cGAN}(G,D)= &
\hspace{-7pt} \mathbb{E}_{(x_a,x_b) \in {\cal X}} [\log D(x_a, H_a, x_b, H_b)] + \\
 & \hspace{-40pt} \mathbb{E}_{(x_a,x_b) \in {\cal X}, z \in {\cal Z}} [\log ( 1 - D(x_a, H_a, \hat{x}, H_b ) )],
\end{array}
\end{equation}

\noindent
where $\hat{x} = G(z, x_a, H_a, H_b)$. 

Previous works on conditional GANs  combine the adversarial loss with either an $L_2$  \cite{DBLP:journals/corr/PathakKDDE16} or an $L_1$-based loss \cite{pix2pix2016,ma2017pose} which is used only for $G$. For instance, the $L_1$ distance computes a pixel-to-pixel difference between the generated and the real image, which, in our case, is:

\begin{equation}
L_1(\hat{x},x_b) =  ||\hat{x} - x_b ||_1.
\end{equation}

\noindent
However,
a well-known problem behind the use of $L_1$ and $L_2$ is the production of blurred images.
We hypothesize that this is also due to the inability of these losses to tolerate small spatial misalignments between $\hat{x}$ and $x_b$. For instance, suppose that 
$\hat{x}$, produced by $G$, is visually plausible and semantically similar to $x_b$, but the texture details on the clothes of the person in the two compared images are not pixel-to-pixel aligned. Both the $L_1$ and the $L_2$ loss will penalize this inexact pixel-level alignment, although not semantically important from the human point of view.
 Note that these misalignments do {\em not} depend on the global deformation between $x_a$ and $x_b$, because  $\hat{x}$ is supposed to have the same pose as $x_b$. In order to alleviate this problem, we propose to use a {\em nearest-neighbour} loss ${\cal L}_{NN}$ based on the following definition of image difference:

\begin{equation}
\label{eq.L-NN}
L_{NN} (\hat{x},x_b) = \sum_{\mathbf{p} \in \hat{x}}  min_{\mathbf{q} \in {\cal N}(\mathbf{p})} || g(\hat{x}(\mathbf{p})) -g(x_b (\mathbf{q})) ||_1,
\end{equation}

\noindent
where ${\cal N}(\mathbf{p})$ is a $n \times n$ local 
neighbourhood of point $\mathbf{p}$
(we use $5 \times 5$ and $3 \times 3$ neighbourhoods for the DeepFashion and the Market-1501 dataset, respectively, see Sec.~\ref{Experiments}).
$g(x (\mathbf{p}))$ is a vectorial representation of a patch around point $\mathbf{p}$ in image $x$, obtained using convolutional filters (see below for more details).
Note that $L_{NN}()$ is not a metrics because it is not symmetric. 
In order to efficiently compute Eq.~\ref{eq.L-NN}, we compare patches in $\hat{x}$ and $x_b$ using their representation
($g()$) in a  
convolutional map of an externally trained network.
In more detail, we use VGG-19 \cite{DBLP:journals/corr/SimonyanZ14a}, trained on ImageNet  and, specifically, its second convolutional layer (called $conv_{1\_2}$). The first two convolutional maps in VGG-19 ($conv_{1\_1}$ and  $conv_{1\_2}$) are both obtained using a convolutional stride equal to 1. For this reason, the feature map ($C_x$) of an image $x$ in  $conv_{1\_2}$ has the same resolution of the original image $x$. Exploiting this fact, we compute the nearest-neighbour field directly on  $conv_{1\_2}$, without losing spatial precision. 
Hence, we define: $g(x (\mathbf{p})) = C_x(\mathbf{p})$, which corresponds to the vector of all the channel values of $C_x$ with respect to the spatial position $\mathbf{p}$. $C_x(\mathbf{p})$ has a receptive field of $5 \times 5$ in $x$, thus effectively representing a patch of dimension $5 \times 5$ using a cascade of two convolutional filters. Using $C_x$, Eq.~\ref{eq.L-NN} becomes:

\begin{equation}
\label{eq.L-NN-conv}
L_{NN} (\hat{x},x_b) = \sum_{\mathbf{p} \in \hat{x}}  min_{\mathbf{q} \in {\cal N}( \mathbf{p})} ||C_{\hat{x}}(\mathbf{p}) - C_{x_b} (\mathbf{q}) ||_1,
\end{equation}

\noindent
 In Sec.~\ref{GPU},
 we show how Eq.~\ref{eq.L-NN-conv} can be efficiently implemented 
  using GPU-based parallel computing.
   The final $L_{NN}$-based loss is:

\begin{equation}
\label{eq.NN-loss}
{\cal L}_{NN}(G) = 
\mathbb{E}_{(x_a,x_b) \in {\cal X}, z \in {\cal Z}} L_{NN}(\hat{x},x_b).
\end{equation}

Combining Eq.~\ref{eq.GAN-loss} and Eq.~\ref{eq.NN-loss} we obtain our  objective:

\begin{equation}
\label{eq.objective}
G^* = \arg \min_G \max_D {\cal L}_{cGAN}(G,D) + \lambda {\cal L}_{NN}(G),
\end{equation}

\noindent
with $\lambda = 0.01$ used
 in all our experiments. The value of $\lambda$ is small because it acts as a normalization factor in  Eq.~\ref{eq.L-NN-conv} with respect to the number of channels in $C_x$ and the number of pixels in  $\hat{x}$ (more details in  
 Sec.~\ref{GPU}).

\section{Implementation details}
\label{sec:impDetails}
We train $G$ and $D$ for 90k iterations, with the Adam optimizer (learning rate: $2 * 10^{-4}$, $\beta_1 = 0.5$, $\beta_2 =0.999$). Following \cite{pix2pix2016} we use instance normalization \cite{DBLP:journals/corr/UlyanovVL16}. 
In the following we denote with: 
(1) $C_m^s$  a convolution-ReLU layer with $m$ filters and stride $s$,
(2)  $CN_m^s$  the same as $C_m^s$ with instance normalization before ReLU and
(3) $CD_m^s$ the same as $CN_m^s$  with the addition of  dropout at rate $50\%$.
Differently from \cite{pix2pix2016}, we use dropout only at training time.
The encoder part of the generator is given by two streams (Fig.~\ref{fig:pipeline}), each of which is composed of the following sequence of layers:

$CN_{64}^1 - CN_{128}^2 - CN_{256}^2 - CN_{512}^2 - CN_{512}^2 - CN_{512}^2$.

\noindent
The decoder part of the generator is given by:

$CD_{512}^2 - CD_{512}^2 - CD_{512}^2 - CN_{256}^2 - CN_{128}^2 - C_{3}^1$.

\noindent
In the last  layer,  ReLU is replaced with $tanh$.

The discriminator architecture is:

$C_{64}^2 - C_{128}^2 - C_{256}^2 - C_{512}^2 - C_{1}^2$,

\noindent
where the  ReLU of the last  layer  is replaced with $sigmoid$.

The generator for the DeepFashion dataset has one additional convolution block ($CN_{512}^2$)
both in the encoder and in the decoder, because images in this dataset  have a  higher resolution.

\section{Experiments}
\label{Experiments}

\paragraph{Datasets}

The person re-identification  Market-1501 dataset \cite{zheng2015scalable} contains 32,668 images of 1,501 persons captured from 6 different surveillance cameras. This dataset is challenging because of the low-resolution images (128$\times$64) and the high diversity in pose, illumination, background and viewpoint. To train our model, we need pairs of images of the same person in two different poses. As this dataset is relatively noisy, we first automatically remove those images in which no human body is detected using the HPE, leading to 263,631 training pairs.
For testing, following  \cite{ma2017pose}, we  randomly select 12,000 pairs. No person is in common between the training and the test split.
 
The DeepFashion dataset ({\em In-shop Clothes Retrieval Benchmark}) \cite{liu2016deepfashion} is composed of 52,712 clothes images, matched each other in order to form  200,000 pairs of identical clothes with two different poses and/or scales of the persons wearing these clothes. 
The images have a resolution of 256$\times$256 pixels. 
Following the training/test split adopted in \cite{ma2017pose}, we create pairs of images, each pair depicting the same person with identical clothes  but in different poses.
After removing those images in which the HPE does not detect any human body, we finally 
 collect 89,262 pairs for training and 12,000 pairs for testing.

\paragraph{Metrics}
Evaluation  in the context of generation tasks is a problem in itself. In our experiments 
 we adopt a redundancy of  metrics and a user study based on human judgments.
Following \cite{ma2017pose}, we use  Structural Similarity (\emph{SSIM}) \cite{wang2004image},  Inception Score (\emph{IS}) \cite{salimans2016improved} and their corresponding masked versions \emph{mask-SSIM} and \emph{mask-IS} \cite{ma2017pose}. The latter are obtained by masking-out the image background and the rationale behind this is that, since no background information of the target image is input to $G$, the network cannot guess what the target background looks like.  
Note that the evaluation masks we use to compute both the mask-IS and the mask-SSIM values do not correspond to the masks ($\{M_h\}$) we use for training. The evaluation masks have been built following the procedure proposed in \cite{ma2017pose} and adopted in that work for both  training and evaluation. Consequently, the mask-based metrics may be biased in favor of their method.
Moreover, we observe that
 the IS metrics  \cite{salimans2016improved}, based on the entropy computed over the classification neurons of an external classifier 
\cite{DBLP:journals/corr/SzegedyVISW15}, 
is not very suitable for
domains with only one object class. 
 For this reason we propose to  use an additional metrics that we call Detection Score (\emph{DS}). Similarly to the classification-based metrics ({\em FCN-score}) used in \cite{pix2pix2016}, DS is based on the detection outcome of
the state-of-the-art object detector SSD \cite{liu2016ssd},
trained on Pascal VOC 07 \cite{pascal-voc-2007} (and not fine-tuned on our datasets).
At testing time, we use the person-class detection scores of SSD 
computed on each generated image $\hat{x}$.
 $DS(\hat{x})$ corresponds to the maximum-score box of SSD on $\hat{x}$ and the final DS value is computed by averaging the scores of all the generated images. In other words, DS measures the confidence of a person detector in the presence of a person in the image. Given the high accuracy of SSD in the challenging Pascal VOC 07 dataset \cite{liu2016ssd}, we believe that it can be used as a good measure of how much realistic (person-like) is a generated image.

Finally, in our tables we also include  the value of each metrics computed using the {\em real} images of the test set. Since these  values are computed on real data, they can be considered as a sort of an upper-bound to the results a generator can obtain. However, these  values are not actual upper bounds in the strict sense: for instance the DS metrics on the real datasets is not 1 because of SSD failures.

\begin{table*}[h!]
\caption{Comparison with the state of the art. $(*)$ These values have been computed 
using the code and the network weights released by Ma et al. \cite{ma2017pose} in order to generate new images.}
\centering
\begin{tabular}{l|ccccc|ccc}
  \hline
  &\multicolumn{5}{c|}{Market-1501}&\multicolumn{3}{c}{DeepFashion}\\
 Model &\emph{SSIM} & \emph{IS}&\emph{mask-SSIM} & \emph{mask-IS} &\emph{DS}& \emph{SSIM} & \emph{IS} &\emph{DS}\\
\hline
Ma et al. \cite{ma2017pose} &\bf$0.253$ & $\bf3.460$ & $0.792$ & $3.435$& $0.39^*$  &$\bf0.762$ & $3.090$ & $0.95^*$  \\
\emph{Ours}&$\bf0.290$ & $3.185$ & $\bf0.805$ & $\bf3.502$& $\bf0.72$ &$0.756$ & $\bf3.439$ & $\bf0.96$ \\
\hline
\emph{Real-Data}&$1.00$ & $3.86$ & $1.00$ & $3.36$& $0.74$ &$1.000$ & $3.898$ & $0.98$ \\
\hline
\end{tabular}
\label{tab:result}
\end{table*}

\subsection{Comparison with previous work}
\label{Comparison}

In Tab.~\ref{tab:result} we compare our method with \cite{ma2017pose}. Note that there are no other works to compare with on this task yet. 
The mask-based metrics are not reported in \cite{ma2017pose} for the DeepFashion dataset.
Concerning the DS metrics,
we used the publicly available code and  network weights released by the authors of \cite{ma2017pose} in order to
generate new images according to the common testing protocol  and  ran the SSD detector to get the DS values.

On the Market-1501 dataset our method
 reports the highest performance with all but the IS metrics. Specifically, our  DS values are much higher than those obtained by \cite{ma2017pose}.
Conversely, on the DeepFashion dataset, our approach significantly improves the IS value but returns a slightly lower SSIM value. 

\subsection{User study}
\label{Userstudy}

In order to further compare our method with the state-of-the-art approach \cite{ma2017pose} 
 we implement a user study 
following the protocol of Ma et al. \cite{ma2017pose}. For each dataset, we show
55 real  and 55 generated images in a random order to 30 users for one second.
Differently from Ma et al. \cite{ma2017pose}, who used Amazon Mechanical Turk (AMT),
we used 
 ``expert'' (voluntary) users: PhD students and Post-docs working in Computer Vision and belonging to two different departments.
 We believe that expert users, who are familiar with GAN-like images,
 can more easily 
 distinguish  real  from fake images,  thus confusing our users is potentially a more difficult task for our GAN.
The results\footnote{$R2G$ means $\#$Real images rated as generated / $\#$Real images; 
$G2R$ means $\#$Generated images rated as Real / $\#$Generated images.}  in 
Tab.~\ref{tab:user-study} 
confirm the significant  quality boost of our images with respect to the images produced in
\cite{ma2017pose}. For instance, on the  Market-1501 dataset, the $G2R$ human ``confusion'' is one order of magnitude higher than in \cite{ma2017pose}.

Finally, in Sec.~\ref{app.comparison} we show some example images, directly comparing with \cite{ma2017pose}.  We also show the results obtained 
by training different person re-identification systems  after augmenting the training set with images generated by our  method. These experiments indirectly confirm that the degree of realism and diversity of our images is very significant.

\begin{table}[h!]
\caption{User study ($\%$).
$(*)$ These results are reported in \cite{ma2017pose} and refer to a similar study with AMT users. } 
\centering
\begin{tabular}{l|cc|cc}
  \hline
  &\multicolumn{2}{c|}{Market-1501}&\multicolumn{2}{c}{DeepFashion}\\
 Model &\emph{R2G} & \emph{G2R} & \emph{R2G} & \emph{G2R} \\
\hline
Ma et al. \cite{ma2017pose}   $(*)$        & 11.2       & 5.5        & 9.2       & 14.9 \\
\emph{Ours}             & 22.67     & 50.24  &  12.42   & 24.61 \\
\hline
\end{tabular}
\label{tab:user-study}
\end{table}

\begin{figure}[h]
  \centering
  \setlength\tabcolsep{0.55pt}
\begin{tabular}{cccccccc}
$x_a$ & $P(x_a)$& $P(x_b)$& $x_b$  & \small\emph{Baseline}& \small\emph{DSC} & \small\emph{PercLoss} & \small\emph{Full}\\ 
\includegraphics[width=0.12\columnwidth]{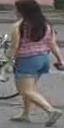}
&\includegraphics[width=0.12\columnwidth]{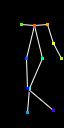} 
&\includegraphics[width=0.12\columnwidth]{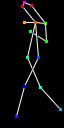}
&\includegraphics[width=0.12\columnwidth]{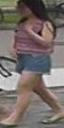}
&\includegraphics[width=0.12\columnwidth]{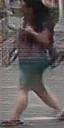}
&\includegraphics[width=0.12\columnwidth]{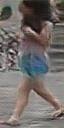}
&\includegraphics[width=0.12\columnwidth]{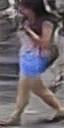}
&\includegraphics[width=0.12\columnwidth]{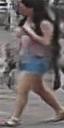}
\\ 
\includegraphics[width=0.12\columnwidth]{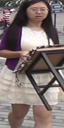}
&\includegraphics[width=0.12\columnwidth]{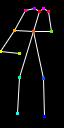} 
&\includegraphics[width=0.12\columnwidth]{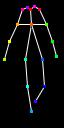}
&\includegraphics[width=0.12\columnwidth]{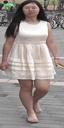}
&\includegraphics[width=0.12\columnwidth]{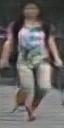}
&\includegraphics[width=0.12\columnwidth]{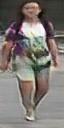}
&\includegraphics[width=0.12\columnwidth]{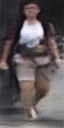}
&\includegraphics[width=0.12\columnwidth]{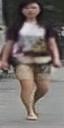}
\\ 
\includegraphics[width=0.12\columnwidth]{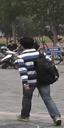}
&\includegraphics[width=0.12\columnwidth]{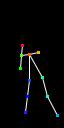} 
&\includegraphics[width=0.12\columnwidth]{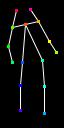}
&\includegraphics[width=0.12\columnwidth]{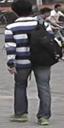}
&\includegraphics[width=0.12\columnwidth]{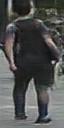}
&\includegraphics[width=0.12\columnwidth]{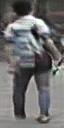}
&\includegraphics[width=0.12\columnwidth]{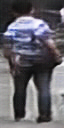}
&\includegraphics[width=0.12\columnwidth]{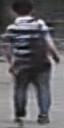}
\\ 
\includegraphics[width=0.12\columnwidth]{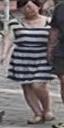}
&\includegraphics[width=0.12\columnwidth]{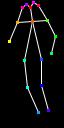} 
&\includegraphics[width=0.12\columnwidth]{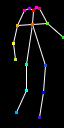}
&\includegraphics[width=0.12\columnwidth]{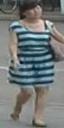}
&\includegraphics[width=0.12\columnwidth]{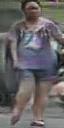}
&\includegraphics[width=0.12\columnwidth]{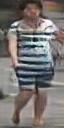}
&\includegraphics[width=0.12\columnwidth]{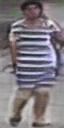}
&\includegraphics[width=0.12\columnwidth]{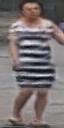}
\\ 
\includegraphics[width=0.12\columnwidth]{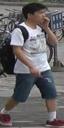}
&\includegraphics[width=0.12\columnwidth]{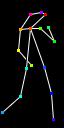} 
&\includegraphics[width=0.12\columnwidth]{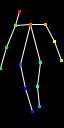}
&\includegraphics[width=0.12\columnwidth]{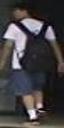}
&\includegraphics[width=0.12\columnwidth]{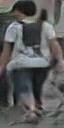}
&\includegraphics[width=0.12\columnwidth]{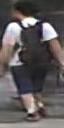}
&\includegraphics[width=0.12\columnwidth]{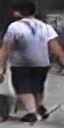}
&\includegraphics[width=0.12\columnwidth]{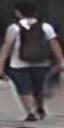}
\end{tabular}
  \caption{Qualitative results on the Market-1501 dataset. Columns 1, 2 and 3 represent the input of our model.
  We plot $P(\cdot)$  as a skeleton for the sake of clarity, but actually no joint-connectivity relation is exploited in our approach.
  Column 4 corresponds to the ground truth. The last four columns show the output of our approach with respect to different baselines.}
\label{fig:ablationMarket}
\end{figure}
\begin{figure}[h]
  \centering
  \setlength\tabcolsep{0.55pt}
\begin{tabular}{cccccccc}
$x_a$ & $P(x_a)$& $P(x_b)$& $x_b$  & \small\emph{Baseline}& \small\emph{DSC} &  \small\emph{PercLoss} & \small\emph{Full}\\ 
\includegraphics[width=0.12\columnwidth]{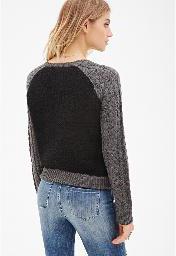}
&\includegraphics[width=0.12\columnwidth]{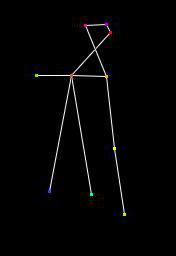} 
&\includegraphics[width=0.12\columnwidth]{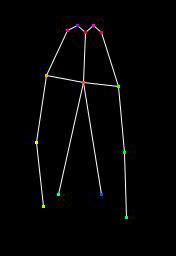}
&\includegraphics[width=0.12\columnwidth]{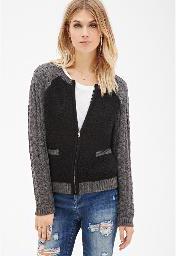}
&\includegraphics[width=0.12\columnwidth]{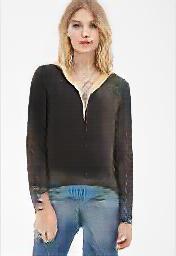}
&\includegraphics[width=0.12\columnwidth]{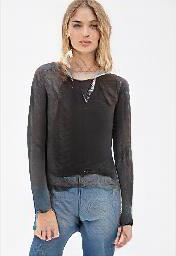}
&\includegraphics[width=0.12\columnwidth]{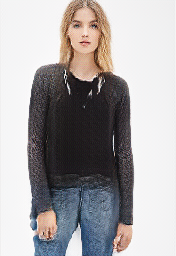}
&\includegraphics[width=0.12\columnwidth]{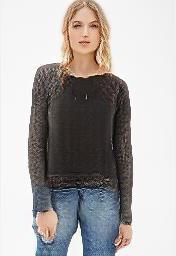}
\\ 
\includegraphics[width=0.12\columnwidth]{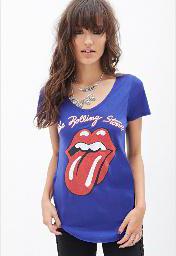}
&\includegraphics[width=0.12\columnwidth]{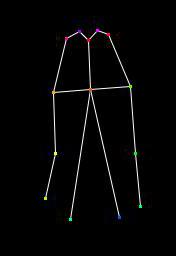} 
&\includegraphics[width=0.12\columnwidth]{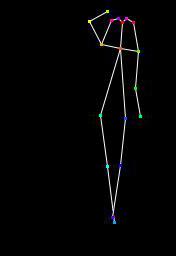}
&\includegraphics[width=0.12\columnwidth]{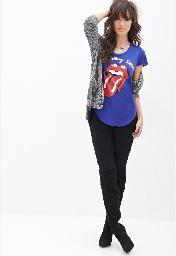}
&\includegraphics[width=0.12\columnwidth]{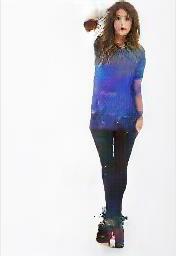}
&\includegraphics[width=0.12\columnwidth]{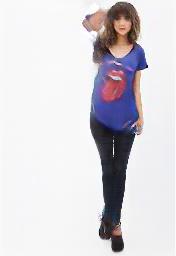}
&\includegraphics[width=0.12\columnwidth]{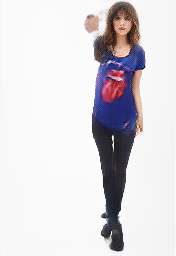}
&\includegraphics[width=0.12\columnwidth]{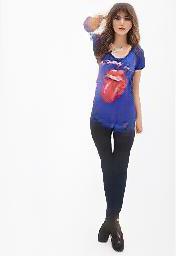}
\\ 
\includegraphics[width=0.12\columnwidth]{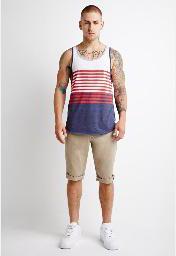}
&\includegraphics[width=0.12\columnwidth]{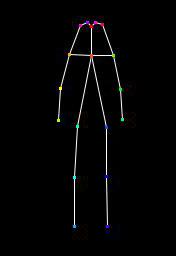} 
&\includegraphics[width=0.12\columnwidth]{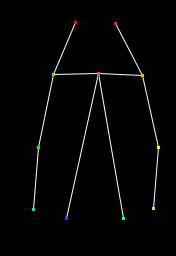}
&\includegraphics[width=0.12\columnwidth]{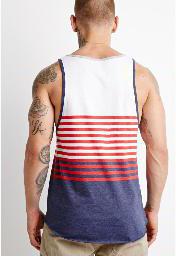}
&\includegraphics[width=0.12\columnwidth]{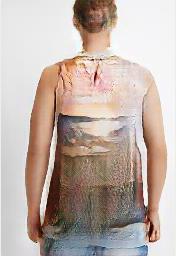}
&\includegraphics[width=0.12\columnwidth]{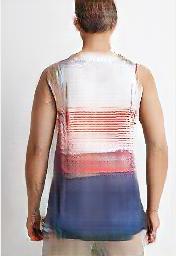}
&\includegraphics[width=0.12\columnwidth]{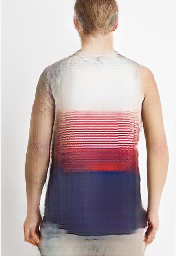}
&\includegraphics[width=0.12\columnwidth]{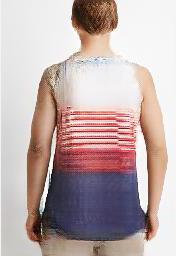}
\\ 
\includegraphics[width=0.12\columnwidth]{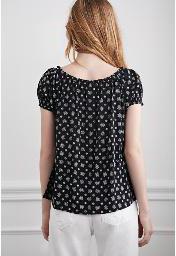}
&\includegraphics[width=0.12\columnwidth]{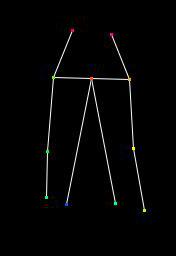} 
&\includegraphics[width=0.12\columnwidth]{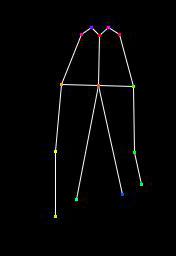}
&\includegraphics[width=0.12\columnwidth]{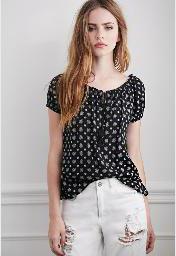}
&\includegraphics[width=0.12\columnwidth]{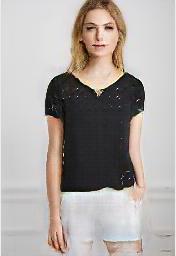}
&\includegraphics[width=0.12\columnwidth]{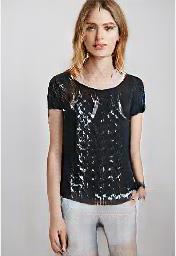}
&\includegraphics[width=0.12\columnwidth]{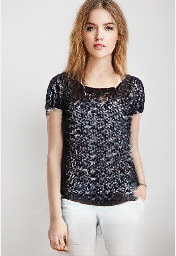}
&\includegraphics[width=0.12\columnwidth]{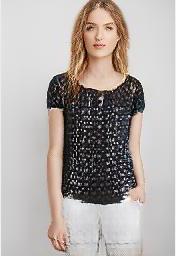}
\\
\includegraphics[width=0.12\columnwidth]{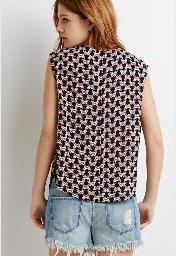}
&\includegraphics[width=0.12\columnwidth]{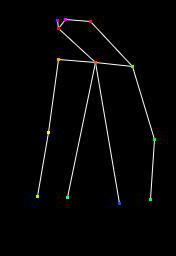} 
&\includegraphics[width=0.12\columnwidth]{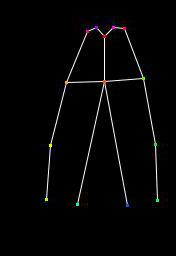}
&\includegraphics[width=0.12\columnwidth]{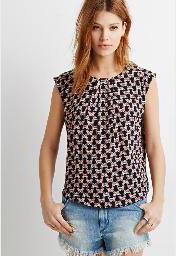}
&\includegraphics[width=0.12\columnwidth]{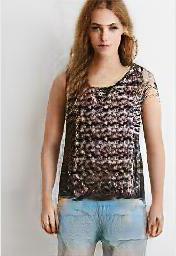}
&\includegraphics[width=0.12\columnwidth]{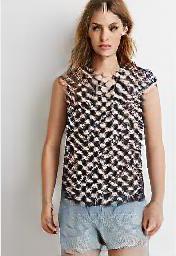}
&\includegraphics[width=0.12\columnwidth]{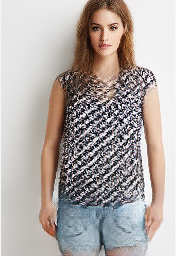}
&\includegraphics[width=0.12\columnwidth]{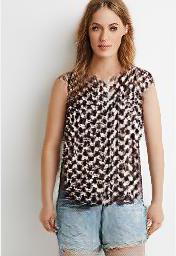}
\\
\includegraphics[width=0.12\columnwidth]{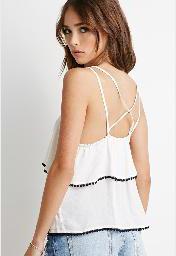}
&\includegraphics[width=0.12\columnwidth]{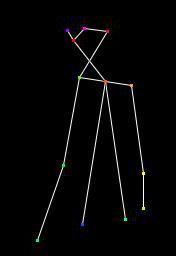} 
&\includegraphics[width=0.12\columnwidth]{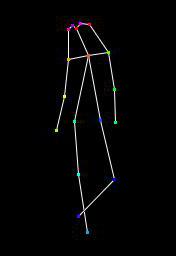}
&\includegraphics[width=0.12\columnwidth]{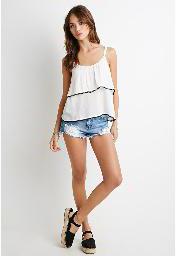}
&\includegraphics[width=0.12\columnwidth]{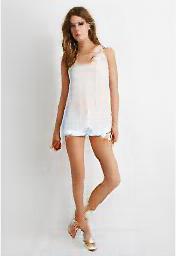}
&\includegraphics[width=0.12\columnwidth]{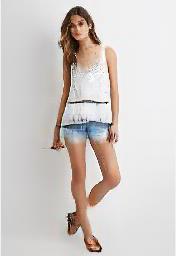}
&\includegraphics[width=0.12\columnwidth]{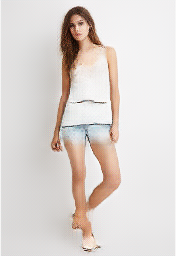}
&\includegraphics[width=0.12\columnwidth]{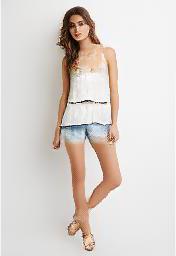}
\end{tabular}
  \caption{Qualitative results on the DeepFashion dataset with respect to different baselines. 
  Some images  have been cropped  
  for visualization purposes.}
\label{fig:ablationFashion}
\end{figure}

\begin{table*}[h]
\caption{Quantitative ablation study on the Market-1501 and the DeepFashion dataset.}
\centering
\begin{tabular}{l|ccccc|cc}
  \hline
  &\multicolumn{5}{c|}{Market-1501}&\multicolumn{2}{c}{DeepFashion}\\
Model &\emph{SSIM} & \emph{IS}&\emph{mask-SSIM} & \emph{mask-IS} &\emph{DS}&\emph{SSIM} & \emph{IS}\\
\hline
\emph{Baseline}&$0.256$ & $3.188$ & $0.784$ & $3.580$& $0.595$ &$0.754$ & $3.351$  \\
\emph{DSC}&$0.272$ & $\bf3.442$ & $0.796$ & $\bf3.666$& $0.629$ &$0.754$ & $3.352$  \\
\emph{PercLoss} & $0.276$ & $3.342$ & $0.788$ & $3.519$ & $0.603$ & $0.744$ & $3.271$  \\
\emph{Full}&$\bf0.290$ & $3.185$ & $\bf0.805$ & $3.502$ & $\bf0.720$ &$\bf0.756$ & $\bf3.439$ \\
\hline
\emph{Real-Data}&$1.00$ & $3.86$ & $1.00$ & $3.36$& $0.74$ &$1.000$ & $3.898$ \\
\hline
\end{tabular}
\label{tab:ablation}
\end{table*}

\subsection{Ablation study and qualitative analysis}
\label{Ablation}
In this section we present  an ablation study to clarify the impact of each part of our proposal on the final performance. We first describe the compared methods, obtained by ``amputating'' important parts of the full-pipeline presented in Sec.~\ref{architectures}-\ref{Training}. The discriminator architecture is the same for all the methods. 

\begin{itemize}
\item \emph{Baseline}:  We use the standard U-Net architecture  \cite{pix2pix2016} {\em without} deformable skip connections. The inputs of $G$ and $D$ and the way pose information is represented (see the definition of tensor $H$ in Sec.~\ref{architectures}) is the same as in 
the full-pipeline. However, in $G$, $x_a$, $H_a$ and $H_b$ are
 concatenated at the input layer. Hence,
 the encoder of $G$ is  composed of only one stream, whose architecture is the same as the two streams  described in Sec.\ref{sec:impDetails}. 

\item \emph{DSC}: $G$ is implemented as described in Sec.~\ref{architectures}, introducing our Deformable Skip Connections (DSC). Both in DSC and in Baseline, training is performed using an $L_1$ loss together with the adversarial loss.

\item
\emph{PercLoss}: This is DSC in which the $L_1$ loss is replaced with the Perceptual loss proposed in \cite{DBLP:conf/eccv/JohnsonAF16}. 
This loss is computed using the layer $conv_{2\_1}$ of 
\cite{DBLP:journals/corr/SimonyanZ14a}, chosen to have a receptive field the closest possible to ${\cal N}( \mathbf{p})$ in Eq.~\ref{eq.L-NN-conv}, and computing the element-to-element difference in this layer  {\em without} nearest neighbor search.

\item \emph{Full}: This is the full-pipeline whose results are reported in Tab.~\ref{tab:result}, and in which we use the proposed ${\cal L}_{NN}$ loss (see Sec.~\ref{Training}).
\end{itemize}

In Tab.~\ref{tab:ablation} we report a quantitative evaluation on the Market-1501 and on the DeepFashion dataset with respect to the four different versions of our approach. 
In most of the cases, there is a   progressive improvement  from  Baseline to  DSC to Full. Moreover, Full usually obtains better results than 
PercLoss. These improvements are particularly evident looking at the DS metrics, which we believe it is a strong evidence that the generated images are realistic.  
 DS values on the DeepFashion dataset are omitted because they are all close to the value $\sim0.96$. 

In Fig.~\ref{fig:ablationMarket} and Fig.~\ref{fig:ablationFashion} we show some qualitative results.
These figures show the progressive improvement through the four baselines which is quantitatively presented above.
In fact, while pose information is usually well generated by all the methods, 
the texture generated by Baseline often does not correspond to the texture in $x_a$ or is blurred. In same cases, the improvement of Full with respect to Baseline is quite drastic, such as the drawing on the shirt of the girl in the second row of Fig.~\ref{fig:ablationFashion} or 
the stripes on the clothes of the persons in the third and in the fourth row of Fig.~\ref{fig:ablationMarket}. Further examples are shown in the Appendix.

\section{Conclusions}
In this paper we presented a GAN-based approach for image generation of persons conditioned on the appearance and  the pose. We introduced two novelties: deformable skip connections and nearest-neighbour loss. The first is used to solve common problems in U-Net based generators when dealing with deformable objects. 
The second novelty is used to alleviate a different type of misalignment between the generated image and the ground-truth image.

Our experiments, based on both automatic evaluation metrics and human judgments,
show that the proposed method is able to outperform previous work on this task.
Despite the proposed method was tested on the specific task of human-generation, only few assumptions are used which refer to the human body and we believe that our proposal can be easily extended to address other deformable-object generation tasks.

\subsection*{Acknowledgements}
We want to thank the NVIDIA Corporation for the  donation  of  the  GPUs  used  in this project.

{\small
\bibliographystyle{ieee}
\bibliography{egbib}
}
\clearpage

\appendix
\section*{Appendix}
\label{Appendix}

In this Appendix we report some additional implementation details and we show other quantitative and qualitative results.
Specifically, in Sec.~\ref{GPU} we explain how Eq.~\ref{eq.L-NN-conv} can be efficiently implemented using GPU-based parallel computing, while in Sec.~\ref{symmetry} we show how the human-body symmetry can be exploited in case of missed limb detections.
In Sec.~\ref{PersonRe-ID} we train state-of-the-art Person Re-IDentification (Re-ID) systems using a combination of real and generated data, which, on the one hand, shows how our images can be effectively used to boost the performance of discriminative methods and, on the other hand, indirectly shows that our generated images are realistic and diverse.
In Sec.~\ref{app.comparison} we show a direct (qualitative) comparison of our method with the approach presented in \cite{ma2017pose} and 
 in Sec.~\ref{Other} we show other images generated by our method, including some failure cases.
Note that some of the images in the DeepFashion dataset have been manually cropped (after the automatic generation) to improve the overall visualization quality.

\section{Nearest-neighbour loss implementation}
\label{GPU}

Our proposed nearest-neighbour loss 
is based on the  definition of $L_{NN} (\hat{x}, x_b)$ given in Eq.~\ref{eq.L-NN-conv}.
In that equation, for each point $\mathbf{p}$ in $\hat{x}$, the ``most similar'' (in the $C_x$-based feature space) point $\mathbf{q}$ in $x_b$ needs to be searched for in a $n \times n$ neighborhood of $\mathbf{p}$.
This operation may be quite time consuming if implemented using sequential computing (i.e., using a ``\texttt{for-loop}'').
We show here how this computation can be 
sped-up by exploiting GPU-based parallel computing in which different tensors are processed simultaneously. 

Given  $C_{x_b}$, we  compute $n^2$ 
shifted versions of $C_{x_b}$: $\{C_{x_b}^{(i, j)}\}$, where $(i,j)$ is a translation offset ranging in a relative $n \times n$ neighborhood
($i,j \in \{ - \frac{n-1}{2}, ..., + \frac{n-1}{2} \}$)  
and $C_{x_b}^{(i, j)}$
is filled with the value $+\infty$ in the borders. Using this translated versions of  $C_{x_b}$,
we compute $n^2$ corresponding difference tensors $\{D^{(i, j)}\}$, where:

\begin{equation}
D^{(i, j)} = |C_{\hat{x}} - C_{x_b}^{(i, j)}| 
\end{equation}

\noindent
and the difference is computed element-wise. $D^{(i, j)} (\mathbf{p})$ contains the channel-by-channel  absolute difference between 
 $C_{\hat{x}} (\mathbf{p})$ and $C_{x_b}(\mathbf{p} + (i, j))$.
 Then, for each $D^{(i, j)}$, we sum all the channel-based differences obtaining:  

\begin{equation} 
\label{eq.S}
 S^{(i, j)} = \sum_c D^{(i, j)}(c),
\end{equation} 

\noindent 
 where $c$ ranges over all the channels and the sum is performed  
pointwise. 
$S^{(i, j)}$ is a matrix of scalar values, each value representing the $L_1$ norm of the difference between a point $\mathbf{p}$ in $C_{\hat{x}}$ and a corresponding point $\mathbf{p} + (i, j)$ in  $C_{x_b}$:

\begin{equation}
S^{(i, j)}(\mathbf{p}) = ||C_{\hat{x}} (\mathbf{p}) - C_{x_b}(\mathbf{p} + (i, j)) ||_1.
\end{equation}

For each point $\mathbf{p}$, we can now compute its best match in a local neighbourhood of  $C_{x_b}$ simply using:

\begin{equation}
\label{eq.M}
M(\mathbf{p}) =   min_{(i, j)} S^{(i, j)}(\mathbf{p}).
\end{equation}

Finally,
Eq.~\ref{eq.L-NN-conv} becomes:

\begin{equation}
\label{eq.M-sum}
L_{NN}(\hat{x},x_b) =   \sum_{\mathbf{p}} M(\mathbf{p}).
\end{equation}

Since we do not normalize  Eq.~\ref{eq.S} by the number of  channels nor   Eq.~\ref{eq.M-sum} by the number of pixels,
the final value $L_{NN}(\hat{x},x_b)$  is usually very high. For this reason we use a small value $\lambda = 0.01$ in Eq.~\ref{eq.objective} when weighting ${\cal L}_{NN}$ with respect to ${\cal L}_{cGAN}$.

\section{Exploiting the human-body symmetry}
\label{symmetry}

As mentioned in Sec.~\ref{skip-connections}, we decompose the human body in 10 rigid sub-parts: the head, the torso and 8 limbs (left/right upper/lower arm, etc.). When one of the joints corresponding to one of these body-parts has not been detected by the HPE, the corresponding region and affine transformation are not computed and  the region-mask is filled with 0. This can happen because of either that region is not visible in the input image or because of false-detections of the HPE. 

However, when the missing region involves a limb (e.g., the right-upper arm) whose symmetric body part has been detected (e.g., the left-upper arm), we can ``copy'' information from the ``twin'' part.
In more detail, suppose for instance that the region corresponding to the right-upper arm in the conditioning image is $R_{rua}^a$ and this region is empty because of one of the above reasons. Moreover, suppose that $R_{rua}^b$ is the corresponding (non-empty) region in $x_b$ and that $R_{lua}^a$
is the (non-empty)  left-upper arm region in $x_a$. We simply set: $R_{rua}^a := R_{lua}^a$ and we compute $f_{rua}$ as usual, using the (now, no more empty) region $R_{rua}^a$ together with  $R_{rua}^b$.

\section{Improving person Re-ID via data-augmentation}
\label{PersonRe-ID}

The goal of this section is to show that the synthetic images generated with our proposed approach can be used to train discriminative methods. Specifically, we use Re-ID approaches whose task is to recognize a human person in different poses and viewpoints. The typical application  of a Re-ID system is a video-surveillance  scenario  in which
 images of the same person, grabbed by 
cameras mounted in different  locations, need to be matched to  each other. Due to the low-resolution of the cameras, person re-identification is usually based on the colours and the texture of the clothes  \cite{DBLP:journals/corr/ZhengYH16}. This makes our method particularly suited to automatically populate a Re-ID training dataset by generating images of a given person with identical clothes but in different viewpoints/poses. 

In our   experiments we use   
Re-ID methods taken from \cite{DBLP:journals/corr/ZhengYH16,DBLP:journals/tomccap/ZhengZY18} and we refer the reader to those papers for details about the involved approaches.
We employ the Market-1501 dataset that is designed for Re-ID method benchmarking.
For each image of the Market-1501 training dataset ($\mathcal{T}$), we randomly select 10 target poses, generating 10 corresponding images using our approach.
Note that: (1) Each generated image is labeled with the {\em identity} of the conditioning image, (2) 
The target {\em pose} can be extracted from an individual different from the person depicted in the conditioning image
(this is different from the other experiments shown here and in the main paper).
Adding the generated images to $\mathcal{T}$ we obtain an augmented training set $\mathcal{A}$. 
In Tab.~\ref{tab:Re-ID} 
we report the results obtained using either $\mathcal{T}$ (standard procedure) or $\mathcal{A}$
for training different Re-ID systems.
 The strong performance boost, orthogonal to different Re-ID methods,
shows that our generative approach can be effectively used 
 for synthesizing training samples.
It also indirectly shows that the generated images are sufficiently realistic and different from the real images contained in $\mathcal{T}$.

\begin{table*}[h]
\caption{Accuracy of Re-ID methods  on the Market-1501 test set ($\%$)}
\centering
\begin{tabular}{l|cc|cc}
  \hline
  &\multicolumn{2}{c|}{Standard training set ($\mathcal{T}$)}&\multicolumn{2}{c}{Augmented training set ($\mathcal{A}$)}\\
 Model &\emph{Rank 1} & \emph{mAP} & \emph{Rank 1} & \emph{mAP} \\
\hline
IDE + Euclidean \cite{DBLP:journals/corr/ZhengYH16}    &  73.9  &  48.8 &  \bf 78.5 & \bf 55.9  \\
IDE + XQDA \cite{DBLP:journals/corr/ZhengYH16}           &  73.2  &  50.9  & \bf  77.8 & \bf 57.9  \\
IDE + KISSME \cite{DBLP:journals/corr/ZhengYH16}        &  75.1  &  51.5 &  \bf 79.5 & \bf 58.1  \\
Discriminative Embedding \cite{DBLP:journals/tomccap/ZhengZY18}  &  78.3  &  55.5  &  \bf 80.6 & \bf 61.3  \\
\hline
\end{tabular}
\label{tab:Re-ID}
\end{table*}

\section{Comparison with previous work}
\label{app.comparison}

In this section we directly compare our method with the results generated by Ma et al. \cite{ma2017pose}. The comparison is based on  the pairs conditioning image-target pose used in \cite{ma2017pose}, for which we show both the results obtained by Ma et al. \cite{ma2017pose} and ours.

\begin{figure*}[h]
  \centering
  \setlength\tabcolsep{1.0pt}
\begin{tabular}{cccc}
$x_a$ & $x_b$ & \small\emph{Full (ours)}& Ma et al. \cite{ma2017pose}\\ 
\includegraphics[width=0.11\linewidth]{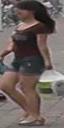}
&\includegraphics[width=0.11\linewidth]{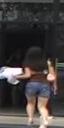}
&\includegraphics[width=0.11\linewidth]{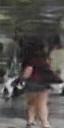}
&\includegraphics[width=0.11\linewidth]{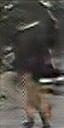}
\\
\includegraphics[width=0.11\linewidth]{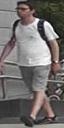}
&\includegraphics[width=0.11\linewidth]{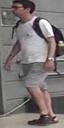}
&\includegraphics[width=0.11\linewidth]{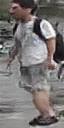}
&\includegraphics[width=0.11\linewidth]{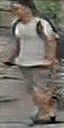}
\\
\includegraphics[width=0.11\linewidth]{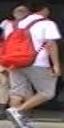}
&\includegraphics[width=0.11\linewidth]{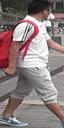}
&\includegraphics[width=0.11\linewidth]{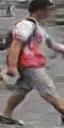}
&\includegraphics[width=0.11\linewidth]{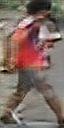}
\\
\includegraphics[width=0.11\linewidth]{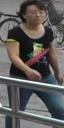}
&\includegraphics[width=0.11\linewidth]{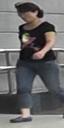}
&\includegraphics[width=0.11\linewidth]{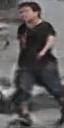}
&\includegraphics[width=0.11\linewidth]{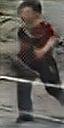}
\\
\includegraphics[width=0.11\linewidth]{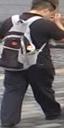}
&\includegraphics[width=0.11\linewidth]{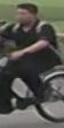}
&\includegraphics[width=0.11\linewidth]{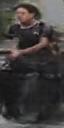}
&\includegraphics[width=0.11\linewidth]{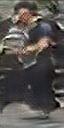}
\end{tabular}
  \caption{A qualitative comparison on the Market-1501 dataset between our approach and the results obtained by Ma et al. \cite{ma2017pose}. Columns 1 and 2  show the conditioning  and the target image, respectively, which are used as reference by both models. Columns 3 and 4 respectively show the images generated by our full-pipeline and by the full-pipeline presented in \cite{ma2017pose}.}
\label{fig:comparison-Market-1}
\end{figure*}

\begin{figure*}[h]
  \centering
  \setlength\tabcolsep{1.0pt}
\begin{tabular}{cccc}
$x_a$ & $x_b$  & \small\emph{Full (ours)}& Ma et al. \cite{ma2017pose}\\ 
\includegraphics[width=0.11\linewidth]{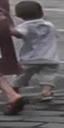}
&\includegraphics[width=0.11\linewidth]{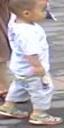}
&\includegraphics[width=0.11\linewidth]{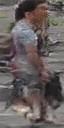}
&\includegraphics[width=0.11\linewidth]{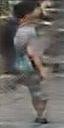}
\\
\includegraphics[width=0.11\linewidth]{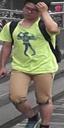}
&\includegraphics[width=0.11\linewidth]{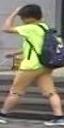}
&\includegraphics[width=0.11\linewidth]{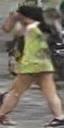}
&\includegraphics[width=0.11\linewidth]{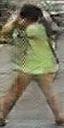}
\\
\includegraphics[width=0.11\linewidth]{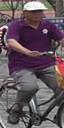}
&\includegraphics[width=0.11\linewidth]{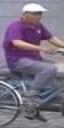}
&\includegraphics[width=0.11\linewidth]{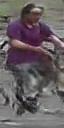}
&\includegraphics[width=0.11\linewidth]{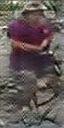}
\\
\includegraphics[width=0.11\linewidth]{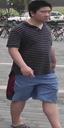}
&\includegraphics[width=0.11\linewidth]{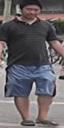}
&\includegraphics[width=0.11\linewidth]{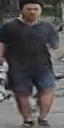}
&\includegraphics[width=0.11\linewidth]{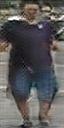}
\\
\includegraphics[width=0.11\linewidth]{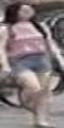}
&\includegraphics[width=0.11\linewidth]{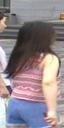}
&\includegraphics[width=0.11\linewidth]{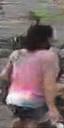}
&\includegraphics[width=0.11\linewidth]{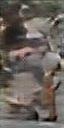}
\end{tabular}
  \caption{More qualitative comparison on the Market-1501 dataset between our approach and the results obtained by Ma et al. \cite{ma2017pose}.}
\label{fig:comparison-Market-2}
\end{figure*}

Figs.~\ref{fig:comparison-Market-1}-\ref{fig:comparison-Market-2} show the results on the Market-1501 dataset. Comparing the images generated by our full-pipeline with the corresponding images generated by the full-pipeline presented in \cite{ma2017pose}, most of the times our results are more realistic, sharper and with local details (e.g., the clothes  texture or the face characteristics) more similar to the details of the conditioning image. 
For instance, in the first and the last row of Fig.~\ref{fig:comparison-Market-1} and in the last row of Fig.~\ref{fig:comparison-Market-2}, our results show  human-like images, while the method proposed in 
\cite{ma2017pose} produced images which can hardly be  recognized as humans.

\begin{figure*}[h]
  \centering
  \setlength\tabcolsep{1.0pt}
\begin{tabular}{cccc}
$x_a$ & $x_b$ & \small\emph{Full (ours)}& Ma et al. \cite{ma2017pose}\\ 
\includegraphics[width=0.12\linewidth]{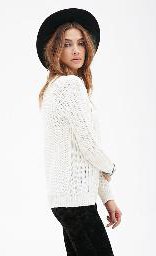}
&\includegraphics[width=0.12\linewidth]{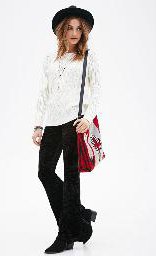}
&\includegraphics[width=0.12\linewidth]{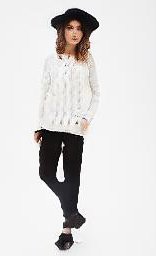}
&\includegraphics[width=0.12\linewidth]{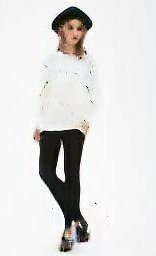}
\\
\includegraphics[width=0.12\linewidth]{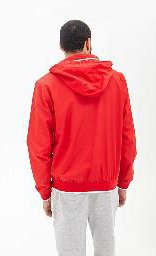}
&\includegraphics[width=0.12\linewidth]{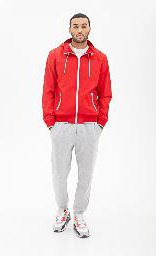}
&\includegraphics[width=0.12\linewidth]{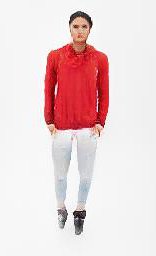}
&\includegraphics[width=0.12\linewidth]{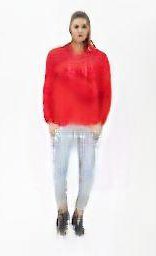}
\\
\includegraphics[width=0.12\linewidth]{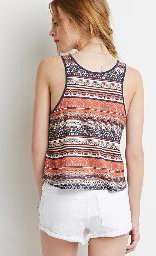}
&\includegraphics[width=0.12\linewidth]{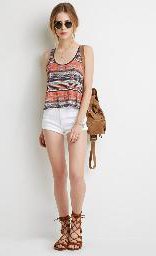}
&\includegraphics[width=0.12\linewidth]{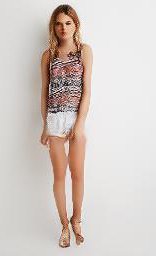}
&\includegraphics[width=0.12\linewidth]{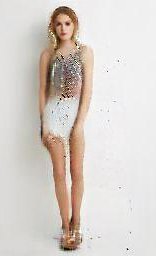}
\\
\includegraphics[width=0.12\linewidth]{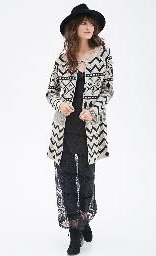}
&\includegraphics[width=0.12\linewidth]{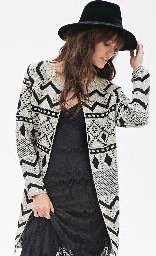}
&\includegraphics[width=0.12\linewidth]{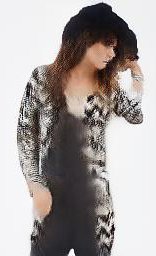}
&\includegraphics[width=0.12\linewidth]{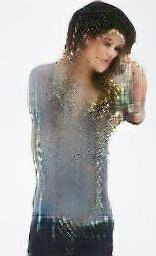}
\\
\includegraphics[width=0.12\linewidth]{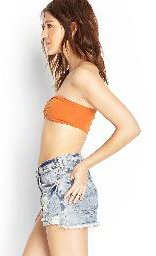}
&\includegraphics[width=0.12\linewidth]{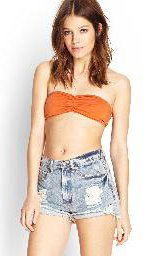}
&\includegraphics[width=0.12\linewidth]{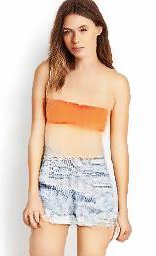}
&\includegraphics[width=0.12\linewidth]{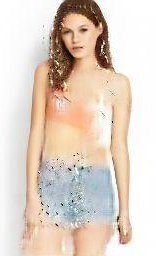}\\
\includegraphics[width=0.12\linewidth]{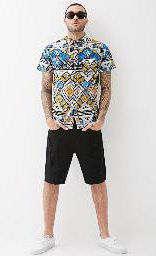}
&\includegraphics[width=0.12\linewidth]{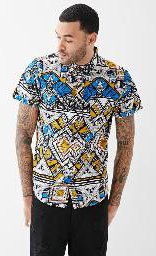}
&\includegraphics[width=0.12\linewidth]{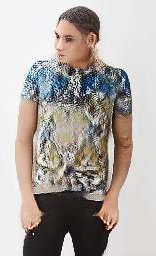}
&\includegraphics[width=0.12\linewidth]{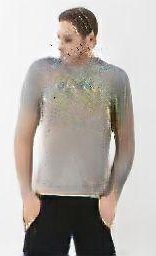}
\end{tabular}
  \caption{A qualitative comparison on the DeepFashion dataset between our approach and the results obtained by Ma et al. \cite{ma2017pose}.}
\label{fig:comparison-Fashion-1}
\end{figure*}

\begin{figure*}[h]
  \centering
  \setlength\tabcolsep{1.0pt}
\begin{tabular}{cccc}
$x_a$ & $x_b$  & \small\emph{Full (ours)}& Ma et al. \cite{ma2017pose}\\ 
\includegraphics[width=0.12\linewidth]{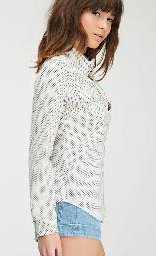}
&\includegraphics[width=0.12\linewidth]{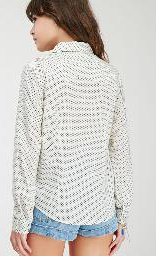}
&\includegraphics[width=0.12\linewidth]{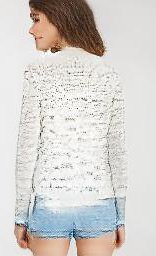}
&\includegraphics[width=0.12\linewidth]{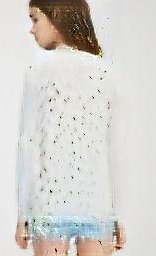}
\\
\includegraphics[width=0.12\linewidth]{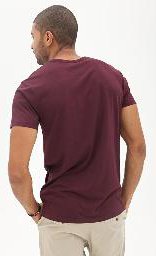}
&\includegraphics[width=0.12\linewidth]{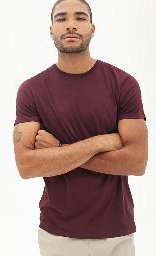}
&\includegraphics[width=0.12\linewidth]{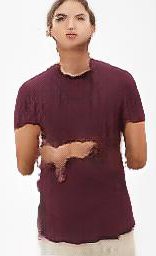}
&\includegraphics[width=0.12\linewidth]{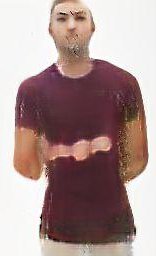}
\\
\includegraphics[width=0.12\linewidth]{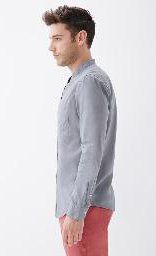}
&\includegraphics[width=0.12\linewidth]{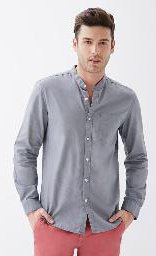}
&\includegraphics[width=0.12\linewidth]{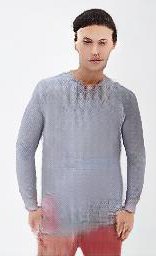}
&\includegraphics[width=0.12\linewidth]{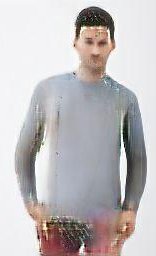}
\\
\includegraphics[width=0.12\linewidth]{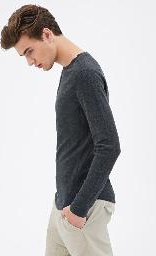}
&\includegraphics[width=0.12\linewidth]{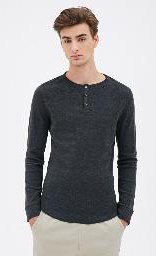}
&\includegraphics[width=0.12\linewidth]{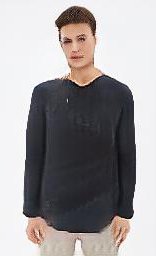}
&\includegraphics[width=0.12\linewidth]{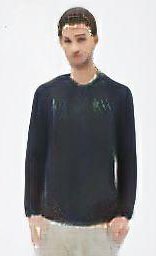}
\\
\includegraphics[width=0.12\linewidth]{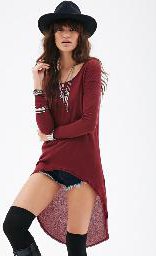}
&\includegraphics[width=0.12\linewidth]{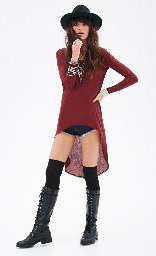}
&\includegraphics[width=0.12\linewidth]{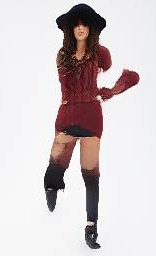}
&\includegraphics[width=0.12\linewidth]{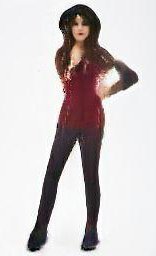}
\\
\includegraphics[width=0.12\linewidth]{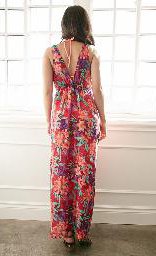}
&\includegraphics[width=0.12\linewidth]{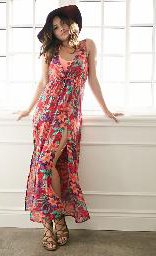}
&\includegraphics[width=0.12\linewidth]{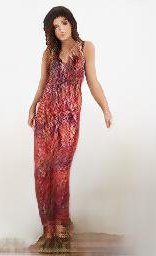}
&\includegraphics[width=0.12\linewidth]{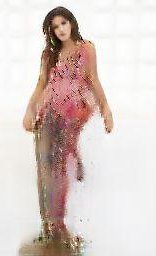}
\end{tabular}
  \caption{More qualitative comparison on the DeepFashion dataset between our approach and the results obtained by Ma et al. \cite{ma2017pose}.}
\label{fig:comparison-Fashion-2}
\end{figure*}

Figs.~\ref{fig:comparison-Fashion-1}-\ref{fig:comparison-Fashion-2} show the results on the DeepFashion dataset. Also in this case, comparing our results with \cite{ma2017pose}, most of the times ours  look more realistic or closer to the details of the conditioning image.  For instance, the second row of Fig.~\ref{fig:comparison-Fashion-1} shows a male face, while the approach proposed in \cite{ma2017pose} produced a female face (note that the DeepFashion dataset is strongly biased toward female subjects \cite{ma2017pose}). Most of the times, the clothes texture in our case is closer to that depicted in the conditioning image (e.g., see rows 1, 3, 4, 5 and 6 in 
Fig.~\ref{fig:comparison-Fashion-1} 
and rows 1 and 6 in  Fig.~\ref{fig:comparison-Fashion-2}). In row 5 of Fig.~\ref{fig:comparison-Fashion-2} the method proposed in 
\cite{ma2017pose} produced an image with a pose closer to the target; however it wrongly generated pants while our approach correctly generated the  appearance of the legs according to the appearance contained in the conditioning image.

We believe that this qualitative comparison {\em using the pairs selected in \cite{ma2017pose}},
 shows that the combination of the proposed deformable skip-connections and the nearest-neighbour loss produced the desired effect to ``capture'' and transfer the correct local details from the conditioning image to the generated image.
 Transferring local information while simultaneously taking into account the global pose deformation is a difficult task which can more hardly be implemented using ``standard''
U-Net based generators as those adopted in \cite{ma2017pose}.

\section{Other qualitative results}
\label{Other}

\begin{figure*}[h]
  \centering
  \setlength\tabcolsep{0.5pt}
\begin{tabular}{cccccccc}
$x_a$ & $P(x_a)$& $P(x_b)$& $x_b$  & \small\emph{Baseline (ours)}& \small\emph{DSC (ours)} & \small\emph{PercLoss (ours)} & \small\emph{Full (ours)}\\ \includegraphics[width=0.11\linewidth]{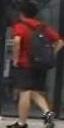}
&\includegraphics[width=0.11\linewidth]{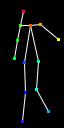} 
&\includegraphics[width=0.11\linewidth]{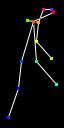}
&\includegraphics[width=0.11\linewidth]{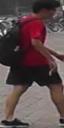}
&\includegraphics[width=0.11\linewidth]{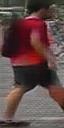}
&\includegraphics[width=0.11\linewidth]{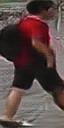}
&\includegraphics[width=0.11\linewidth]{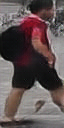}
&\includegraphics[width=0.11\linewidth]{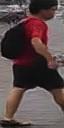}
\\
\includegraphics[width=0.11\linewidth]{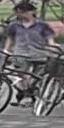}
&\includegraphics[width=0.11\linewidth]{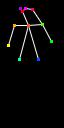} 
&\includegraphics[width=0.11\linewidth]{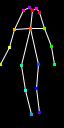}
&\includegraphics[width=0.11\linewidth]{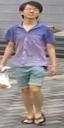}
&\includegraphics[width=0.11\linewidth]{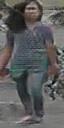}
&\includegraphics[width=0.11\linewidth]{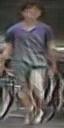}
&\includegraphics[width=0.11\linewidth]{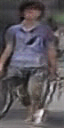}
&\includegraphics[width=0.11\linewidth]{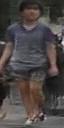}
\\
\includegraphics[width=0.11\linewidth]{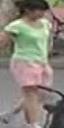}
&\includegraphics[width=0.11\linewidth]{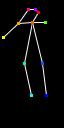} 
&\includegraphics[width=0.11\linewidth]{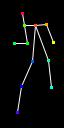}
&\includegraphics[width=0.11\linewidth]{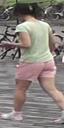}
&\includegraphics[width=0.11\linewidth]{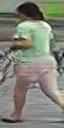}
&\includegraphics[width=0.11\linewidth]{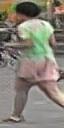}
&\includegraphics[width=0.11\linewidth]{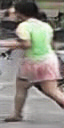}
&\includegraphics[width=0.11\linewidth]{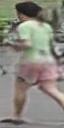}
\\
\includegraphics[width=0.11\linewidth]{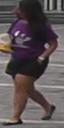}
&\includegraphics[width=0.11\linewidth]{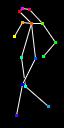} 
&\includegraphics[width=0.11\linewidth]{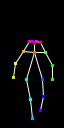}
&\includegraphics[width=0.11\linewidth]{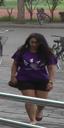}
&\includegraphics[width=0.11\linewidth]{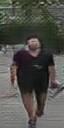}
&\includegraphics[width=0.11\linewidth]{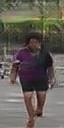}
&\includegraphics[width=0.11\linewidth]{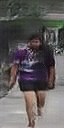}
&\includegraphics[width=0.11\linewidth]{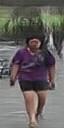}
\\
\includegraphics[width=0.11\linewidth]{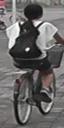}
&\includegraphics[width=0.11\linewidth]{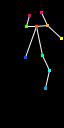} 
&\includegraphics[width=0.11\linewidth]{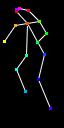}
&\includegraphics[width=0.11\linewidth]{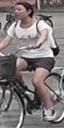}
&\includegraphics[width=0.11\linewidth]{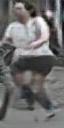}
&\includegraphics[width=0.11\linewidth]{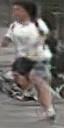}
&\includegraphics[width=0.11\linewidth]{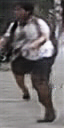}
&\includegraphics[width=0.11\linewidth]{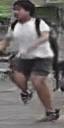}
\end{tabular}

  \caption{Other qualitative  results on the Market-1501 dataset.}
\label{fig:ablationMarket-Good}
\end{figure*}

\begin{figure*}[h]
  \centering
  \setlength\tabcolsep{0.5pt}
\begin{tabular}{cccccccc}
$x_a$ & $P(x_a)$& $P(x_b)$& $x_b$  & \small\emph{Baseline (ours)}& \small\emph{DSC (ours)} & \small\emph{PercLoss (ours)} & \small\emph{Full (ours)}\\ \includegraphics[width=0.12\linewidth]{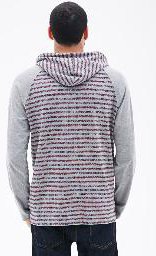}
&\includegraphics[width=0.12\linewidth]{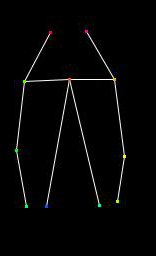} 
&\includegraphics[width=0.12\linewidth]{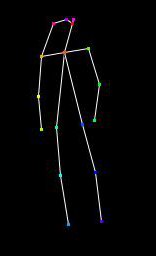}
&\includegraphics[width=0.12\linewidth]{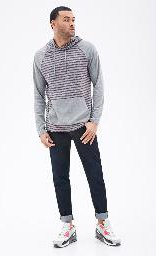}
&\includegraphics[width=0.12\linewidth]{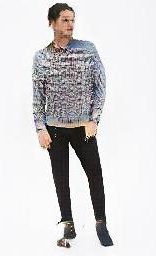}
&\includegraphics[width=0.12\linewidth]{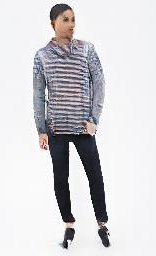}
&\includegraphics[width=0.12\linewidth]{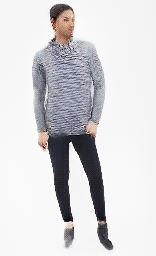}
&\includegraphics[width=0.12\linewidth]{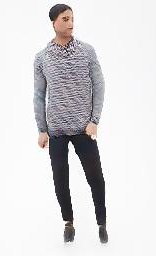}
\\
\includegraphics[width=0.12\linewidth]{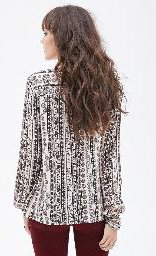}
&\includegraphics[width=0.12\linewidth]{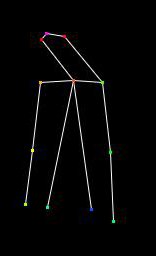} 
&\includegraphics[width=0.12\linewidth]{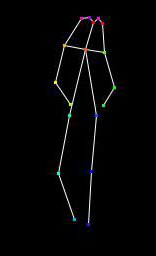}
&\includegraphics[width=0.12\linewidth]{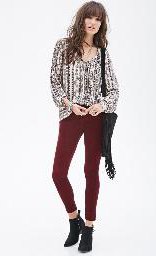}
&\includegraphics[width=0.12\linewidth]{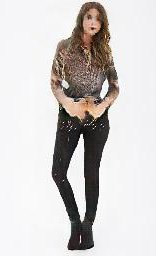}
&\includegraphics[width=0.12\linewidth]{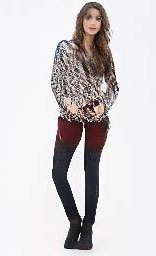}
&\includegraphics[width=0.12\linewidth]{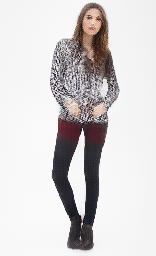}
&\includegraphics[width=0.12\linewidth]{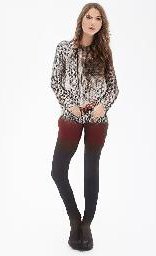}
\\
\includegraphics[width=0.12\linewidth]{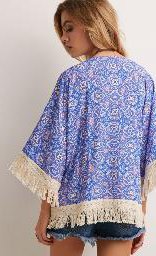}
&\includegraphics[width=0.12\linewidth]{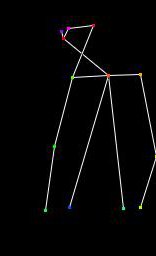} 
&\includegraphics[width=0.12\linewidth]{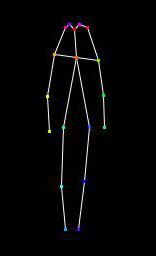}
&\includegraphics[width=0.12\linewidth]{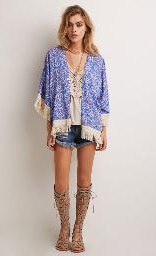}
&\includegraphics[width=0.12\linewidth]{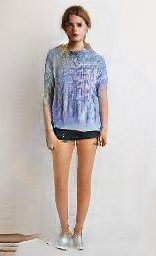}
&\includegraphics[width=0.12\linewidth]{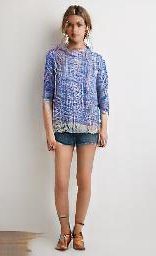}
&\includegraphics[width=0.12\linewidth]{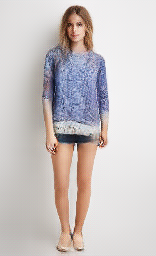}
&\includegraphics[width=0.12\linewidth]{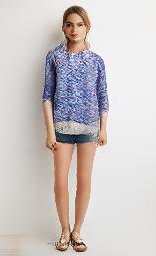}
\\
\includegraphics[width=0.12\linewidth]{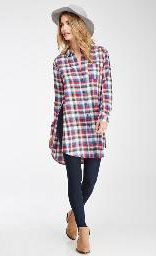}
&\includegraphics[width=0.12\linewidth]{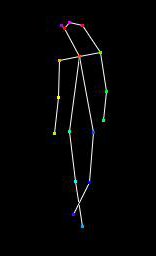} 
&\includegraphics[width=0.12\linewidth]{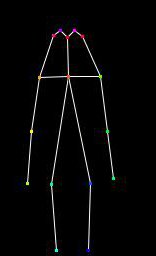}
&\includegraphics[width=0.12\linewidth]{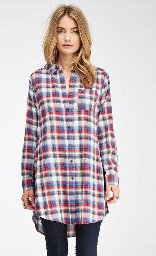}
&\includegraphics[width=0.12\linewidth]{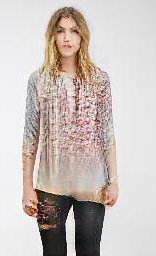}
&\includegraphics[width=0.12\linewidth]{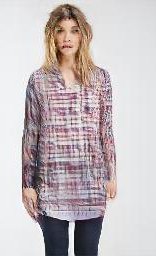}
&\includegraphics[width=0.12\linewidth]{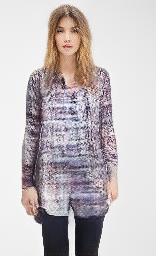}
&\includegraphics[width=0.12\linewidth]{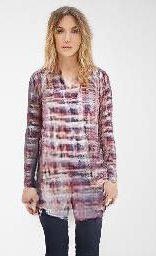}
\\
\includegraphics[width=0.12\linewidth]{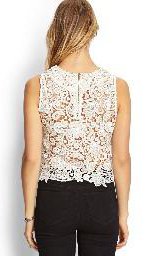}
&\includegraphics[width=0.12\linewidth]{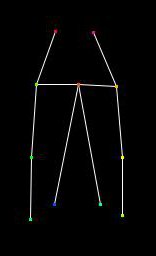} 
&\includegraphics[width=0.12\linewidth]{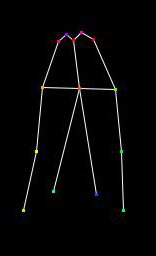}
&\includegraphics[width=0.12\linewidth]{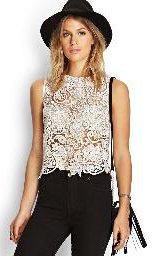}
&\includegraphics[width=0.12\linewidth]{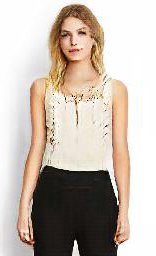}
&\includegraphics[width=0.12\linewidth]{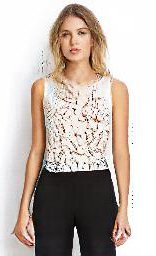}
&\includegraphics[width=0.12\linewidth]{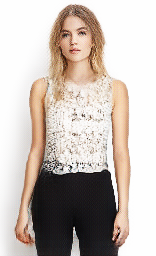}
&\includegraphics[width=0.12\linewidth]{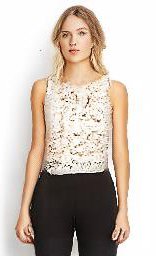}
\end{tabular}
  \caption{Other qualitative results on the DeepFashion dataset.}
\label{fig:ablationFashion-Good}
\end{figure*}

In this section we present  other qualitative results. Fig.~\ref{fig:ablationMarket-Good} and 
Fig.~\ref{fig:ablationFashion-Good}
show some images generated using the Market-1501 dataset and the DeepFashion dataset, respectively. The terminology is the same adopted in Sec.~\ref{Userstudy}. Note that, for the sake of clarity, we used a skeleton-based visualization of $P(\cdot)$ but, as explained in the main paper, only the point-wise joint locations are used in our method to represent pose information (i.e., no joint-connectivity information is used).

Similarly to the results shown in Sec.~\ref{Userstudy}, also these images show that, 
despite the pose-related general structure is sufficiently well generated by all the different versions of our method,
most of the times there is a gradual quality improvement in the detail synthesis from Baseline to DSC to PercLoss to Full.

\begin{figure*}[h]
  \centering
  \setlength\tabcolsep{0.5pt}
\begin{tabular}{cccccccc}
  $x_a$ & $P(x_a)$& $P(x_b)$& $x_b$  & \small\emph{Baseline (ours)}& \small\emph{DSC (ours)} & \small\emph{PercLoss (ours)} & \small\emph{Full (ours)}\\
  \includegraphics[width=0.11\linewidth]{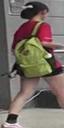}
&\includegraphics[width=0.11\linewidth]{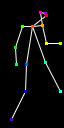} 
&\includegraphics[width=0.11\linewidth]{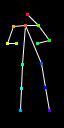}
&\includegraphics[width=0.11\linewidth]{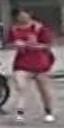}
&\includegraphics[width=0.11\linewidth]{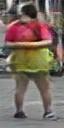}
&\includegraphics[width=0.11\linewidth]{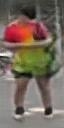}
&\includegraphics[width=0.11\linewidth]{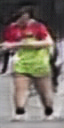}
&\includegraphics[width=0.11\linewidth]{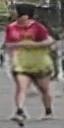}
\\
\includegraphics[width=0.11\linewidth]{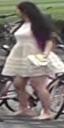}
&\includegraphics[width=0.11\linewidth]{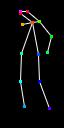} 
&\includegraphics[width=0.11\linewidth]{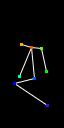}
&\includegraphics[width=0.11\linewidth]{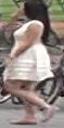}
&\includegraphics[width=0.11\linewidth]{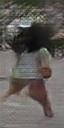}
&\includegraphics[width=0.11\linewidth]{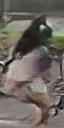}
&\includegraphics[width=0.11\linewidth]{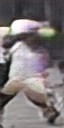}
&\includegraphics[width=0.11\linewidth]{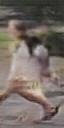}
\\
\includegraphics[width=0.11\linewidth]{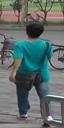}
&\includegraphics[width=0.11\linewidth]{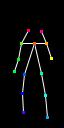} 
&\includegraphics[width=0.11\linewidth]{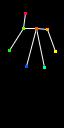}
&\includegraphics[width=0.11\linewidth]{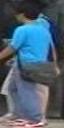}
&\includegraphics[width=0.11\linewidth]{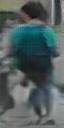}
&\includegraphics[width=0.11\linewidth]{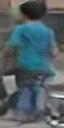}
&\includegraphics[width=0.11\linewidth]{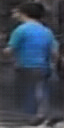}
&\includegraphics[width=0.11\linewidth]{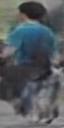}
\\
\includegraphics[width=0.11\linewidth]{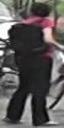}
&\includegraphics[width=0.11\linewidth]{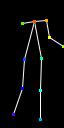} 
&\includegraphics[width=0.11\linewidth]{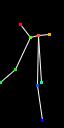}
&\includegraphics[width=0.11\linewidth]{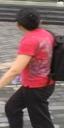}
&\includegraphics[width=0.11\linewidth]{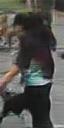}
&\includegraphics[width=0.11\linewidth]{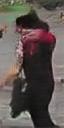}
&\includegraphics[width=0.11\linewidth]{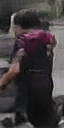}
&\includegraphics[width=0.11\linewidth]{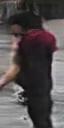}
\\
\includegraphics[width=0.11\linewidth]{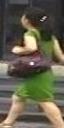}
&\includegraphics[width=0.11\linewidth]{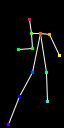} 
&\includegraphics[width=0.11\linewidth]{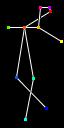}
&\includegraphics[width=0.11\linewidth]{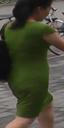}
&\includegraphics[width=0.11\linewidth]{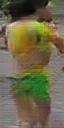}
&\includegraphics[width=0.11\linewidth]{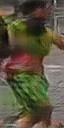}
&\includegraphics[width=0.11\linewidth]{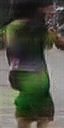}
&\includegraphics[width=0.11\linewidth]{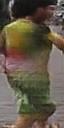}
\end{tabular}
  \caption{Examples of {\em badly} generated images on the Market-1501 dataset. See the text for more details.}
\label{fig:ablationMarket-Fail}
\end{figure*}

\begin{figure*}[h]
  \centering
  \setlength\tabcolsep{0.5pt}
\begin{tabular}{cccccccc}
  $x_a$ & $P(x_a)$& $P(x_b)$& $x_b$  & \small\emph{Baseline (ours)}& \small\emph{DSC (ours)} & \small\emph{PercLoss (ours)} & \small\emph{Full (ours)}\\
  \includegraphics[width=0.12\linewidth]{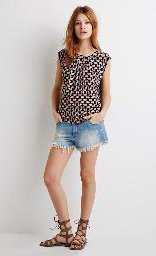}
&\includegraphics[width=0.12\linewidth]{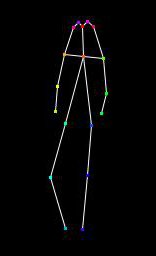} 
&\includegraphics[width=0.12\linewidth]{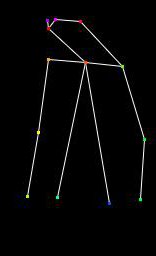}
&\includegraphics[width=0.12\linewidth]{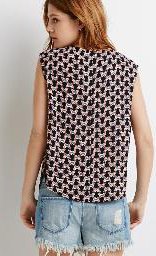}
&\includegraphics[width=0.12\linewidth]{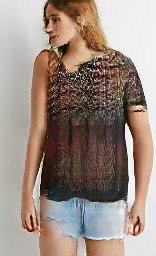}
&\includegraphics[width=0.12\linewidth]{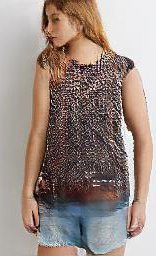}
&\includegraphics[width=0.12\linewidth]{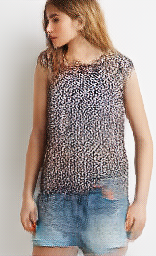}
&\includegraphics[width=0.12\linewidth]{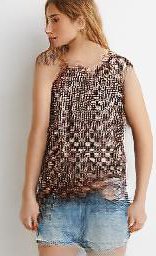}
\\
\includegraphics[width=0.12\linewidth]{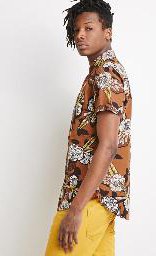}
&\includegraphics[width=0.12\linewidth]{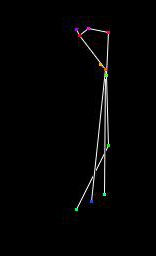} 
&\includegraphics[width=0.12\linewidth]{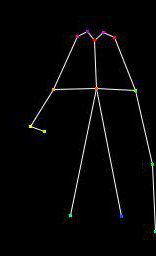}
&\includegraphics[width=0.12\linewidth]{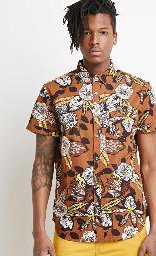}
&\includegraphics[width=0.12\linewidth]{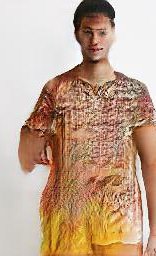}
&\includegraphics[width=0.12\linewidth]{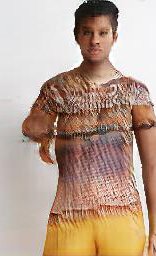}
&\includegraphics[width=0.12\linewidth]{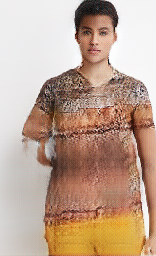}
&\includegraphics[width=0.12\linewidth]{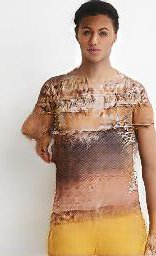}
\\
\includegraphics[width=0.12\linewidth]{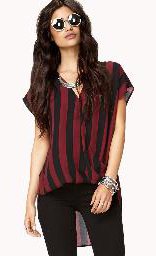}
&\includegraphics[width=0.12\linewidth]{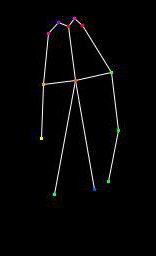} 
&\includegraphics[width=0.12\linewidth]{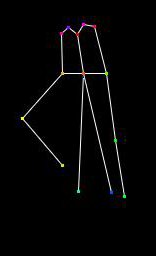}
&\includegraphics[width=0.12\linewidth]{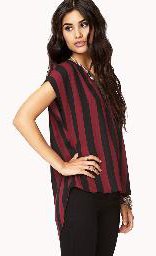}
&\includegraphics[width=0.12\linewidth]{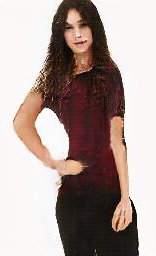}
&\includegraphics[width=0.12\linewidth]{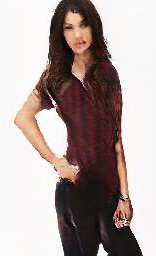}
&\includegraphics[width=0.12\linewidth]{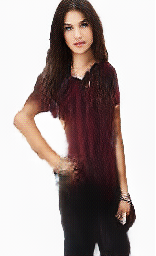}
&\includegraphics[width=0.12\linewidth]{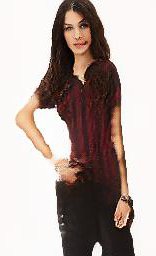}
\\
\includegraphics[width=0.12\linewidth]{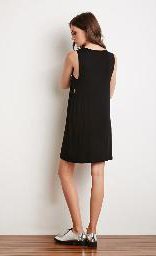}
&\includegraphics[width=0.12\linewidth]{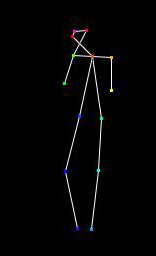} 
&\includegraphics[width=0.12\linewidth]{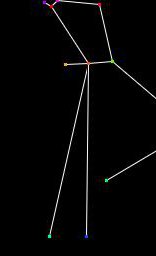}
&\includegraphics[width=0.12\linewidth]{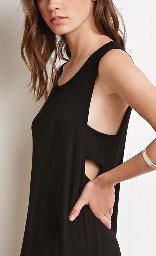}
&\includegraphics[width=0.12\linewidth]{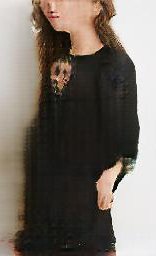}
&\includegraphics[width=0.12\linewidth]{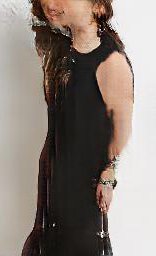}
&\includegraphics[width=0.12\linewidth]{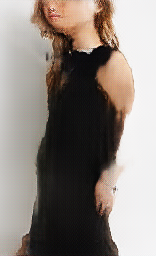}
&\includegraphics[width=0.12\linewidth]{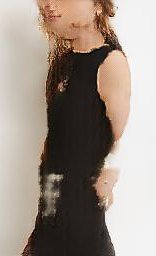}
\\
\includegraphics[width=0.12\linewidth]{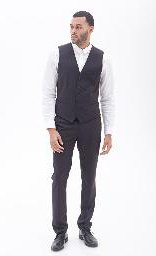}
&\includegraphics[width=0.12\linewidth]{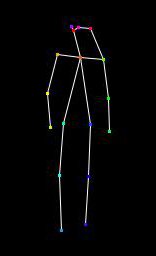} 
&\includegraphics[width=0.12\linewidth]{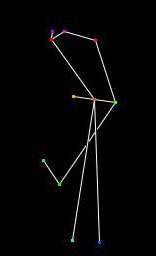}
&\includegraphics[width=0.12\linewidth]{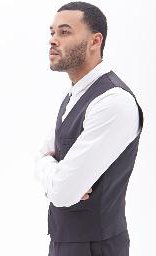}
&\includegraphics[width=0.12\linewidth]{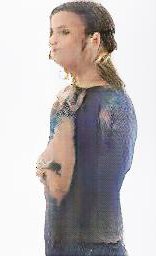}
&\includegraphics[width=0.12\linewidth]{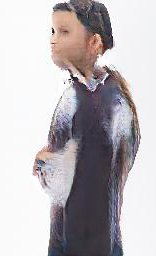}
&\includegraphics[width=0.12\linewidth]{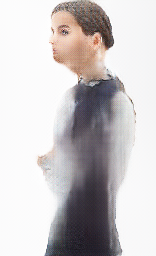}
&\includegraphics[width=0.12\linewidth]{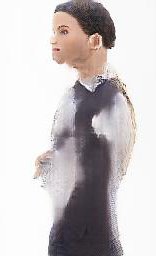}
\end{tabular}
  \caption{Examples of {\em badly} generated images on the DeepFashion dataset.}
\label{fig:ablationFashion-Fail}
\end{figure*}

Finally,  Fig.~\ref{fig:ablationMarket-Fail} and 
Fig.~\ref{fig:ablationFashion-Fail} show some failure cases (badly generated images) of our method on the Market-1501 dataset and the DeepFashion dataset, respectively.
Some common failure  causes are:

\begin{itemize}
\item 
Errors of the HPE \cite{Cao}. For instance, see rows 2, 3 and 4 of  Fig.~\ref{fig:ablationMarket-Fail} or the wrong right-arm localization in row 2 of Fig.~\ref{fig:ablationFashion-Fail}.
\item
Ambiguity of the pose representation. 
For instance,
in  row 3 of Fig.~\ref{fig:ablationFashion-Fail},  the left elbow has been detected in $x_b$ although it is actually hidden behind the body. Since $P(x_b)$ contains only 2D information (no depth or occlusion-related information), there is no way for the system to understand whether the elbow is behind or in front of the body.  In this case our model chose to generate an arm considering that the arm is in front of the body (which corresponds to the most frequent situation in the training dataset). 
 \item 
 Rare poses. For instance, row 1 of Fig.~\ref{fig:ablationFashion-Fail} shows a girl in an unusual  rear view with a sharp 90 degree profile face ($x_b$). The generator by mistake synthesized a neck where it should have ``drawn'' a shoulder. Note that rare poses are a difficult issue also for the method proposed in \cite{ma2017pose}.
 \item 
 Rare object appearance. For instance, the backpack in row 1 of Fig.~\ref{fig:ablationMarket-Fail} is light green, while most of the backpacks contained in the training images of the Market-1501 dataset are dark. Comparing this image with the one generated in the last row of Fig.~\ref{fig:ablationMarket-Good} (where the backpack is black), we see that in Fig.~\ref{fig:ablationMarket-Good} the colour of the shirt of the generated image is not blended with the backpack colour, while in Fig.~\ref{fig:ablationMarket-Fail} it is.
 We presume  that the generator ``understands'' that a dark backpack is an object whose texture should not be transferred to the clothes of the generated image, while it is not able to generalize this knowledge to other backpacks. 
 \item 
 Warping problems. This is an issue related to our specific approach (the deformable skip connections). The texture on the shirt of the conditioning image in row 2 of Fig.~\ref{fig:ablationFashion-Fail} is warped in the generated image. We presume this is due to the fact that in this case the affine transformations  need to largely warp the texture details of the narrow surface of the profile shirt (conditioning image) in order to fit the much wider area of the target frontal pose.
\end{itemize}

\end{document}